\documentclass{article}
\PassOptionsToPackage{numbers, compress}{natbib}
\usepackage[preprint]{configuration/neurips_2023}

\usepackage{xcolor}
\definecolor{darkblue}{rgb}{0.0, 0.0, 0.55}
\definecolor{myblue}{rgb}{0,0.45,0.74}
\definecolor{myred}{rgb}{0.85,0.33,0.1}
\usepackage[hyperindex,breaklinks,colorlinks,citecolor=darkblue]{hyperref} %

\usepackage{stfloats}
\usepackage[utf8]{inputenc}
\usepackage[T1]{fontenc}
\usepackage{booktabs}
\usepackage{amsfonts}       %
\usepackage{nicefrac}       %
\usepackage{microtype}      %
\usepackage[flushleft]{threeparttable}
\usepackage{color}
\usepackage{mathtools}
\usepackage{transparent}

\usepackage{url}
\usepackage{float}
\usepackage{graphicx}
\usepackage{caption}
\usepackage{subcaption}
\usepackage{multirow}
\usepackage{xspace}
\usepackage{natbib}
\usepackage{enumitem}
\usepackage[font=small]{caption}
\usepackage{diagbox}
\usepackage{wrapfig}
\usepackage[toc, page, header]{appendix}
\usepackage{bm}
\usepackage{sidecap}

\usepackage{amsmath,amsthm,amssymb}

\providecommand{\R}{\mathbb{R}} %

\providecommand{\0}{\mathbf{0}}
\providecommand{\1}{\mathbf{1}}

\providecommand{\mm}{\mathbf{m}}

\providecommand{\ww}{\mathbf{w}}
\providecommand{\xx}{\mathbf{x}}

\providecommand{\zz}{\mathbf{z}}

\providecommand{\mH}{\mathbf{H}}
\providecommand{\mI}{\mathbf{I}}

\providecommand{\mK}{\mathbf{K}}
\providecommand{\mL}{\mathbf{L}}

\providecommand{\mX}{\mathbf{X}}
\providecommand{\mY}{\mathbf{Y}}

\newenvironment{talign*}
{\csname align*\endcsname}
{\endalign}
\usepackage{amssymb}

\usepackage{algorithm}
\usepackage[noend]{algpseudocode}

\errorcontextlines\maxdimen

\makeatletter
\newcommand*{\algrule}[1][\algorithmicindent]{\makebox[#1][l]{\hspace*{.5em}\thealgruleextra\vrule height \thealgruleheight depth \thealgruledepth}}%
\newcommand*{\thealgruleextra}{}
\newcommand*{\thealgruleheight}{.75\baselineskip}
\newcommand*{\thealgruledepth}{.25\baselineskip}

\newcount\ALG@printindent@tempcnta
\def\ALG@printindent{%
	\ifnum \theALG@nested>0%
	\ifx\ALG@text\ALG@x@notext%
	\else
		\unskip
		\addvspace{-1pt}%
		\ALG@printindent@tempcnta=1
		\loop
		\algrule[\csname ALG@ind@\the\ALG@printindent@tempcnta\endcsname]%
		\advance \ALG@printindent@tempcnta 1
		\ifnum \ALG@printindent@tempcnta<\numexpr\theALG@nested+1\relax%
			\repeat
		\fi
	\fi
}%
\usepackage{etoolbox}
\patchcmd{\ALG@doentity}{\noindent\hskip\ALG@tlm}{\ALG@printindent}{}{\errmessage{failed to patch}}
\makeatother

\newbox\statebox
\newcommand{\myState}[1]{%
	\setbox\statebox=\vbox{#1}%
	\edef\thealgruleheight{\dimexpr \the\ht\statebox+1pt\relax}%
	\edef\thealgruledepth{\dimexpr \the\dp\statebox+1pt\relax}%
	\ifdim\thealgruleheight<.75\baselineskip
		\def\thealgruleheight{\dimexpr .75\baselineskip+1pt\relax}%
	\fi
	\ifdim\thealgruledepth<.25\baselineskip
		\def\thealgruledepth{\dimexpr .25\baselineskip+1pt\relax}%
	\fi
	\State #1%
	\def\thealgruleheight{\dimexpr .75\baselineskip+1pt\relax}%
	\def\thealgruledepth{\dimexpr .25\baselineskip+1pt\relax}%
}

\usepackage[textwidth=5cm]{todonotes}

\title{Revisiting Implicit Models: Sparsity Trade-offs Capability in Weight-tied Model for Vision Tasks}

\author{%
Haobo Song \\
EPFL, Switzerland \\
\texttt{haobo.song@epfl.ch} \\
\And
Soumajit Majumder \\
Huawei\\
\texttt{soumajit.majumder@huawei.com} \\
\And
Tao Lin\thanks{Corresponding author} \\
Westlake University, P.R.\ China \\
\texttt{lintao@westlake.edu.cn}
}

\begin{document}
\maketitle

\begin{abstract}
	Implicit models such as Deep Equilibrium Models (DEQs) have garnered significant attention in the community for their ability to train infinite layer models with elegant solution-finding procedures and constant memory footprint. However, despite several attempts, these methods are heavily constrained by model inefficiency and optimization instability. Furthermore, fair benchmarking across relevant methods for vision tasks is missing.
	In this work, we revisit the line of implicit models and trace them back to the original weight-tied models. Surprisingly, we observe that weight-tied models are more effective, stable, as well as efficient on vision tasks, compared to the DEQ variants. Through the lens of these simple-yet-clean weight-tied models, we further study the fundamental limits in the model capacity of such models and propose the use of distinct sparse masks to improve the model capacity. Finally, for practitioners, we offer design guidelines regarding the depth, width, and sparsity selection for weight-tied models, and demonstrate the generalizability of our insights to other learning paradigms.
	\looseness=-1
\end{abstract}

\setlength{\parskip}{4pt plus4pt minus0pt}

\section{Introduction}

In recent years, implicit models have gained signification attention in the field of machine learning. Different from classical deep-learning models which rely on explicit computation graphs~\cite{he2016deep}, implicit models characterize their internal mechanism by some pre-specified dynamics.
Classic examples of such implicit models include weight-tied models~\cite{ngiam2010tiled,dehghani2018universal,takase2021lessons}, Neural ODEs~\cite{chen2018neural}, and equilibrium models~\cite{bai2019deep, bai2020multiscale}. These models begin with defining the dynamics of layer iteration and then leverage either black-box ODE solvers~\cite{chen2018neural} or root-finding algorithms~\cite{bai2019deep, bai2020multiscale} to solve the specified dynamics.

Deep Equilibrium Models or DEQs~\cite{bai2019deep,bai2020multiscale} is a prominent implicit model in the research community. The central theme behind DEQ lies in the equilibrium state converged on an infinite-depth network, represented by a fixed point equation. This insight inspires the elegant optimization strategies of DEQ, which empowers the feasibility of achieving a constant memory footprint. Initially introduced for sequence modeling~\cite{bai2019deep}, DEQs were subsequently extended to computer vision applications~\cite{bai2020multiscale}. \looseness=-1

However, achieving stable convergence to a solution in implicit-depth models necessitates substantial tuning~\cite{winston2020monotone}, due to the model's sensitivity to initialization and regularization~\cite{linsley2020stable,bai2021stabilizing,geng2021training,agarwala2022deep}.
To date, an extensive line of research, e.g.~\cite{bai2021stabilizing,geng2021training,agarwala2022deep}, tries to improve upon these known issues of model efficiency or optimization difficulty for sequence models.
Despite these attempts, these issues continue to severely bottleneck the exploration of the potential of such implicit models. \\
Meanwhile, the standard weight-tied models, which inspired DEQ models, and offer both computation \& storage efficiency, remain largely unexamined for vision tasks across~\cite{bai2019deep,bai2020multiscale,bai2021stabilizing,geng2021training,pokle2022deep}.
As our first contribution, we demonstrate that under the same training budget, \emph{weight-tied models offer remarkable prediction and performance efficiency over existing DEQ variants on vision tasks}.

Leveraging the original weight-tied model as a simple proxy on the perspective of feature representation, we identify a fundamental issue, namely restricted model capability (or model expressive power), in most of the implicit models including both weight-tied models and DEQ-like models.
As a remedy, and as our next contribution, we propose \emph{multi-mask weight-tied} to implicitly induce more model capability through diverse sparsity patterns for the tied layers, while enjoying a significantly reduced computational overhead.
Intuitively, storage-free, static, and non-trainable boolean masks are temporally applied to tied layers recursively, resulting in the dissimilar layer structure and thus an increased model capability.
The effectiveness of such a design choice is verified by extensive results.

\textbf{We summarize our contributions below:}
\begin{itemize}[leftmargin=12pt,nosep]
	\item We demonstrate the incredible effectiveness and efficiency of standard weight-tied models over similar implicit models, such as DEQ and its variants.
	      We emphasize heavily that the contribution of this study does not lie in the novelty of the weight-tied model itself; as such a classical idea has occurred in the community with various forms (see~\autoref{sec:related}). Rather, the contribution lies precisely in emphasizing the superior efficiency and effectiveness of such a simple baseline, which should not be omitted for evaluation when proposing advanced implicit model variants.
	\item We leverage the multi-mask weight-tied layer to implicitly induce model capability through the lens of a simple yet clean weight-tied model.
	      The insights therein could further benefit the design of other implicit models in the field, which we leave for future work.
	\item We examine the trade-off between depth, width, and sparsity of the weight-tied layer, through extensive numerical investigations for ResNet- and Vision-Transformer-like models on CIFAR and ImageNet.
	      We provide a clear guideline, as a novel first step, to facilitate the practitioners.
\end{itemize}

\section{Related Work} \label{sec:related}
We provide a compact summary here due to space issues. A complete discussion is in~\autoref{appendix:complete_related_work}.

\paragraph{Implicit models and DEQ variants.}
In recent years, implicit models have garnered widespread attention as they replace explicit layers with a single implicit layer and prescribed internal dynamics, as noted in the works of~\cite{amos2017optnet,chen2018neural,niculae2018sparsemap,wang2019satnet,bai2019deep,bai2020multiscale,bai2021stabilizing,geng2021training,agarwala2022deep}. Among these, DEQs, introduced by~\cite{bai2019deep}, stand as a representative approach in implicit modeling, aimed at discovering the equilibrium of a system to ultimately reach a fixed point equation.
Despite the recent efforts to improve DEQ-like implicit models~\cite{bai2021stabilizing,agarwala2022deep,geng2021training,bai2020multiscale}, most studies largely overlooked the original weight-tied model, despite being simple, effective, and memory inefficient, making the generalizability and practicality of DEQ variants on various use cases to be questioned; our contribution therein.

\paragraph{Weight-tied model.}
Weight-tied models, often referred to as weight-sharing models, are a popular paradigm to achieve parameter-efficient features. These models employ a unified set of weights across diverse layers to largely reduce parameter numbers~\cite{dehghani2018universal,dabre2019recurrent, xia2019tied, lan2020ALBERT,li2021training,takase2021lessons}.
Serving as the key of numerous implicit models, they have been subject to extensive investigation in recent years across a range of applications~\cite{wang2019weight,liu2020comprehensive,yang2018unsupervised,lan2020ALBERT,takase2021lessons,zhang2020deeper,bender2020can,xie2021weight,li2021training}.

While the existing research for the weight-tied model primarily concerns methods for tying diverse layers, they do not encompass the introduction of sparse pruning masks to a shared layer, as in our approach.
Separately, in the context of Neural Architecture Search (NAS), weight-sharing methods are applied to sample distinct neural architectures from a super net with sparse masks to alleviate computational burdens. In this setup, an abundance of architectures can share weights within the same super net and the expensive training procedure can also be reduced to only once.~\cite{zhang2020deeper,bender2020can,xie2021weight}

\paragraph{Model quantization and pruning.}
A line of seminal papers for model quantization~\cite{han2015deep,chen2015compressing} employs the concept of hash functions or quantization to map weights to scalars or codebooks, thereby increasing the compression rate. This approach has been further extended to soft weight sharing~\cite{li2020group,ye2018unified,ullrich2017soft,zhang2018learning}, where the remaining weights are assigned to the most probable clusters. However, this strategy differs from our approach of using sparse masks to enhance capability.

In the realm of model pruning~\cite{wang2023state}.
three main avenues have emerged: i) pruning when initializing, ii) dynamic pruning during training, and iii) pruning after training. The latter two typically involve pruning model weights with extra training or calculation and are thus not efficient.
The method of pruning when initializing first replies on magnitude-based metrics to do pruning~\cite{frankle2018the}. However, several subsequent studies~\cite{su2020sanity,frankle2021pruning,wang2022recent} have ignited a debate, asserting that random masks---randomly sample pruning masks without any prior knowledge---can be just as effective as the earlier ``lottery ticket'' idea~\cite{frankle2018the}.
It's worth noting that most model pruning research traditionally focuses on improving conventional explicit neural networks by refining pruning criteria and proposing advanced optimization strategies or objective functions. As far as our knowledge extends, the introduction of various random masks into a weight-tied model, as presented in our work, is a novel concept.

A related work that bears relevance to our manuscript is~\cite{bai2022parameter}, which has only one physical layer and employs masks atop this fixed layer to generate diverse dense layers.
In particular, this approach utilizes several unique masks to select different sets of values from a random vector (i.e., codebook), thereby creating distinct dense layers.
Nevertheless, it remains distinct from our fundamental idea of utilizing a weight-tied structure to learn a tied weight with deterministic random binary masks, which implicitly imparts model capacity. \looseness=-1

\paragraph{Drouput.}
Dropout, introduced by~\cite{hinton2012improving} serves as a pivotal training technique aimed at mitigating overfitting~\cite{labach2019survey,liu2023dropout}. It achieves this by introducing random modifications to neural network parameters or activations~\cite{wan2013regularization,ba2013adaptive,wang2013fast,kingma2015variational,gal2016dropout}.
While Dropout has found application in compressing neural networks~\cite{molchanov2017variational,neklyudov2017structured,gomez2019learning}, it's important to note that the \emph{stochastic} dropping idea in Dropout is primarily tailored for standard, explicit neural architectures, which stands in contrast to the deterministic masks of our weight-tied models.

\section{Inspecting Implicit Models}
\subsection{Introduction to Deep Equilibrium Model} \label{sec:intro_to_deq}
Deep Equilibrium models (DEQ) are a series of implicit models first introduced by \cite{bai2019deep}. The elegance of such approaches lies in defining the output of the network as the solution to an ``infinite-depth'' fixed point equation. This ingredient enables the use of some root-finding algorithms and therefore avoids the activation storage to achieve a significantly reduced memory footprint.\looseness=-1

\begin{figure*}[!t]
	\centering
	\vspace{-0.5em}
	\includegraphics[width=0.85\textwidth]{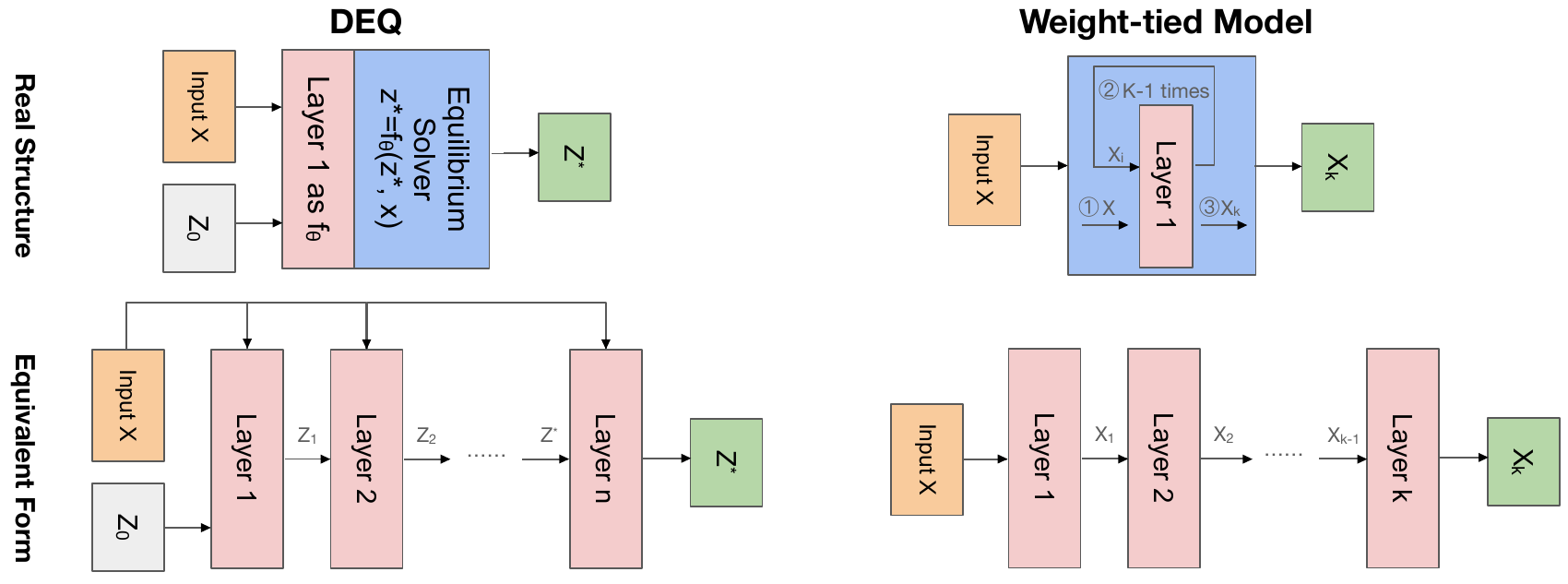}
	\vspace{-0.75em}
	\caption{\small
		\textbf{DEQ and weight-tied model structure schematic diagram.}
		The figure above shows its real structure and the figure below shows its equivalent structure for better understanding.
	}
	\vspace{-1.em}
	\label{fig:structure}
\end{figure*}

\paragraph{Formal definition of DEQ.}
\label{DEQ formulation}
Given a layer $f$ parameterized by $\ww$, the key hypothesis and observation of \cite{bai2019deep} rely on the convergence of the following sequence to a fixed point when increasing the depth/iteration towards infinity:\looseness=-1
\begin{small}
	\begin{align} \label{fixpoint}
		\textstyle
		\lim_{i\to \infty} \zz^{[i]}= \lim_{i \to \infty} f_{\ww}(\zz^{[i]}; \xx) \equiv f_{\ww}(\zz^\star; \xx) = \zz^\star \,,
	\end{align}
\end{small}%
where $\xx \in \R^d$ is the input injection, $\zz^{[0]} := \0 \in \R^d$, and $\zz^{[i+1]}= f_{\ww}(\zz^{[i]}; \xx) \text{ for } i = 0, \cdots, L - 1$.
$\zz^\star$ represents the equilibrium point, or equivalently the root of the equation $g_{\ww}(\zz^{t}; \xx) := f_{\ww}(\zz^{t};\xx) - \zz^{t} = 0$, shown in the left of \autoref{fig:structure}.
A line of attempts~\cite{bai2019deep,bai2021stabilizing,geng2021training} improves the training/optimization phase of these infinite-layer DEQ networks through implicit differentiation and thus enjoys constant memory consumption.
\looseness=-1

\subsection{Tracing Back to the Original Weight-tied Model} \label{sec:intro_to_weight_tied}
Despite the elegance and constant memory cost, it becomes non-trivial to probe other potentials of the implicit layers in DEQ variants, due to the suffered pitfalls of computational inefficiency as well as the optimization instability (see results in~\autoref{appendix:ineffectiveness_DEQ}).
As a result, here we resort to the original weight-tied model given its simplicity and cleanness, and we believe the insights therein could be transformed into other advanced variants of implicit layers (as future work).
\looseness=-1

\paragraph{Formulation.}
The definition of the original weight-tied model largely follows the notations in~\autoref{sec:intro_to_deq}, where a \emph{$K$-depth weight-tied layer} (in the right of~\autoref{fig:structure}) can be modeled explicitly by \looseness=-1
\begin{small}
	\begin{align} \label{eq:weight_tied}
		\xx_{i+1} = f_{\ww}(\xx_{i}) \,,
	\end{align}
\end{small}%
where the index of $i$ in~\eqref{eq:weight_tied} refers to the $i$-th weight-tied layer reusing (representing \emph{$i$-th tied layer}).
Such a design can intuitively reduce the number of parameters by a factor of $K$ but cannot maintain a constant memory footprint like DEQ variants\footnote{
	Note that the pre-training phase of DEQ variants~\cite{bai2019deep,bai2020multiscale,bai2021stabilizing} almost resembles the training procedure of original weight-tied networks, and only differs in the number of training epochs.
	\looseness=-1
}.
\looseness=-1

\subsection{On the Effectiveness of Weight-tied Model} \label{sec:superior_weight_tied_over_deq}
Despite the simplicity and limitation of the weight-tied model, in this subsection, we thoroughly revisit this design choice and assess its efficacy by examining it with various strong competitors on three highly representative neural architectures.
Such an empirical investigation is crucial to the community, given the unknown position of the weight-tied model after years of research on other advanced implicit models, as well as the surprisingly missing comparisons between the weight-tied model and DEQ variants.
\looseness=-1

\begin{figure*}[!t]
	\centering
	\begin{subfigure}{0.315\textwidth}
		\includegraphics[width=1\textwidth]{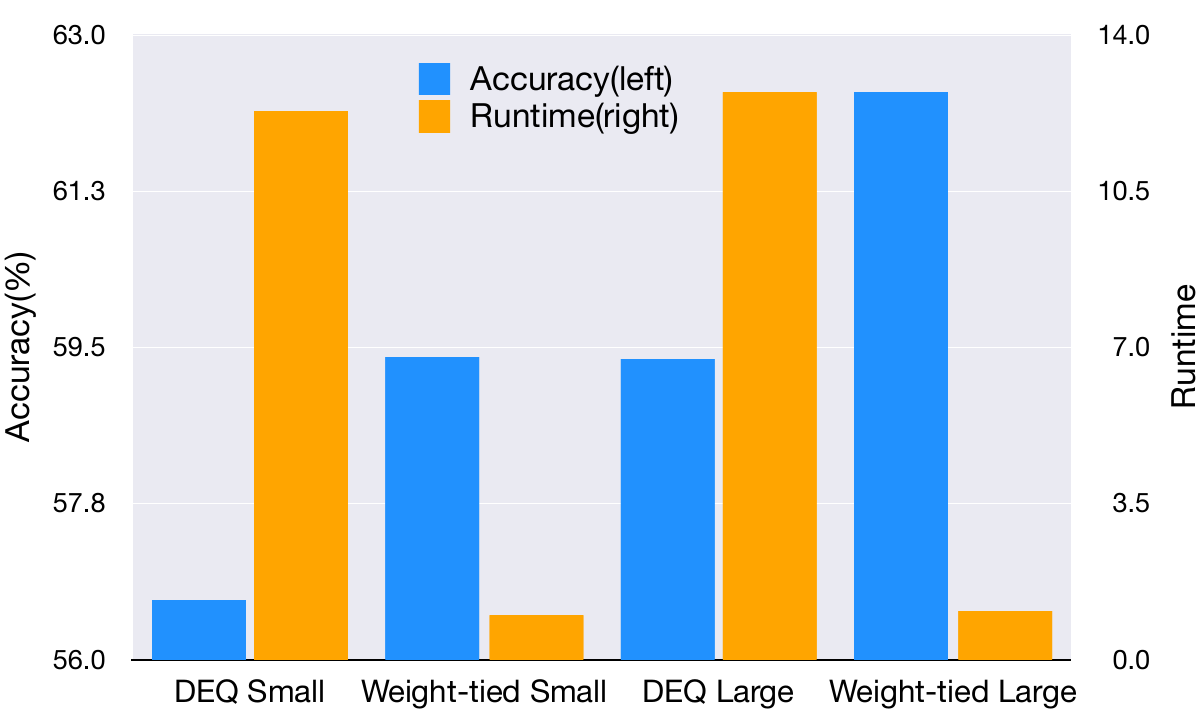}
		\caption{\small MLP}
	\end{subfigure}
	\hfill
	\begin{subfigure}{0.315\textwidth}
		\includegraphics[width=1\textwidth]{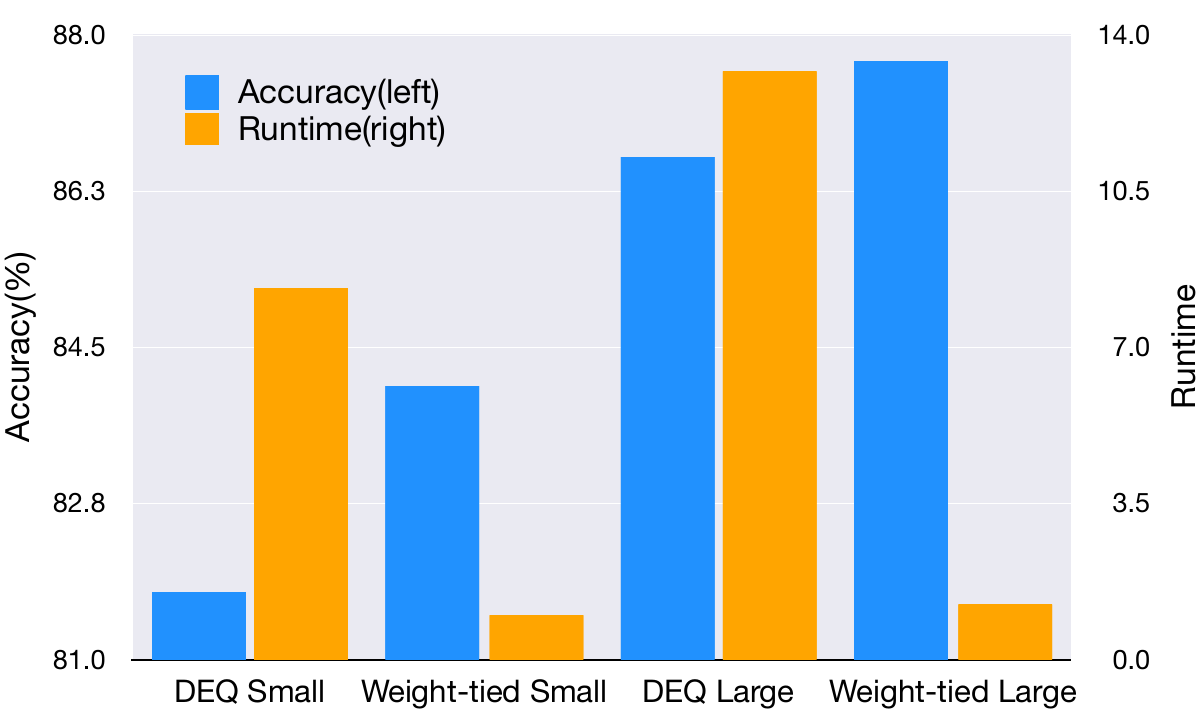}
		\caption{\small ResNet}
	\end{subfigure}
	\hfill
	\begin{subfigure}{0.315\textwidth}
		\includegraphics[width=1\textwidth]{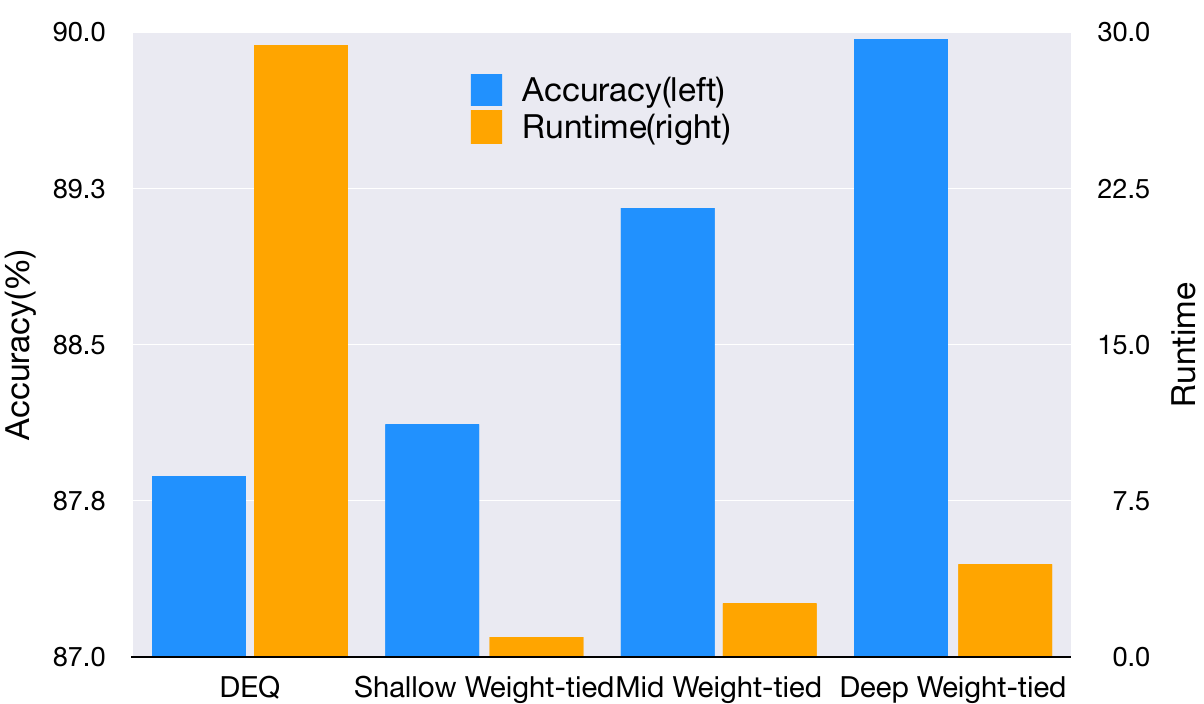}
		\caption{\small CCT}
	\end{subfigure}
	\caption{\small
		\textbf{Weight-tied can surpass DEQ in both computational cost and performance over different structures.}
		We make weight-tied and DEQ comparisons (accuracy v.s.\ runtime) under MLP, ResNet, and CCT structures for the task of CIFAR-10 image classification (w/ data augmentation).
		Runtime refers to the training time used per batch.
		The fastest weight-tied runtime in each structure is chosen as unit time (1x), and the other model's runtime is calculated based on unit time.
		All experiments are allowed sufficient training time and are repeated thrice. The comparison between weight-tied and DEQ is fair, more details are provided in the Appendix.
		The dramatic gap in the runtime cannot be addressed by~\cite{geng2021training}, which could accelerate the training by $1.7 \times$.
	}
	\vspace{-1.em}
	\label{fig:cifar10_various_nn_archs_acc_vs_runtime}
\end{figure*}

\paragraph{Evaluation setup.}
For the sake of simplicity and fair comparison, we transform existing neural architectures into both weight-tied and DEQ models.
Similar to the treatments in DEQ variants~\cite{bai2019deep,bai2020multiscale,bai2021stabilizing,geng2021training} to guard the performance, such transformed networks (for both weight-tied and DEQ models) include a small number of not-tied layers at the bottom and top layer, while the majority of the model is constructed through the tied layer.
The remaining not-tied layers in DEQ and weight-tied model are identical and constitute a very low proportion of the whole model.

We elaborate on the configuration of three considered neural architectures for weight-tied/DEQ models; other training strategies are detailed in~\autoref{appendix:exp_details}.
Note that there exists no difference between the weight-tied model and the DEQ model from the view of parameter space, though the former needs to specify the depth of the tied layer.
\begin{enumerate}[nosep,leftmargin=12pt]
	\item \textbf{MLP}: This MLP only contains linear layers, where the shared part comprises two weight-tied layers.
	      We vary the model capacity, termed as \emph{small} and \emph{large}, by doubling the model width. \looseness=-1
	\item \textbf{ResNet}: ResNet~\cite{he2016deep} is designed similarly as the single-stream model presented in \cite{bai2020multiscale}.
	      The weight-tied version changes the DEQ module(a BasicBlock) in the single-stream model to the weight-tied module and contains four weight-tied layers. Similar notations of \emph{small} and \emph{large} as MLP are used. \looseness=-1
	\item \textbf{CCT}: The neural architecture of CCT (Compact Convolution Transformer~\cite{hassani2021escaping}), considers \emph{shallow}, \emph{medium}, and \emph{deep} weight-tied models, comprising of 3, 5, and 7 weight-tied modules, respectively.
	      We transform its encoder into weight-tied and DEQ structures. \looseness=-1
\end{enumerate}

\paragraph{Observations.}
\autoref{fig:cifar10_various_nn_archs_acc_vs_runtime} illustrates a thorough comparison of accuracy and runtime cost between the DEQ model and the weight-tied model for the task of CIFAR-10 classification.
All CIFAR-10 experiments in this paper are equipped with standard techniques like basic normalization, random cropping, and horizontal flipping.
\emph{In all three structures examined, the weight-tied model demonstrates approximately 2\% higher accuracy and a reduction in runtime ranging from $5 \times$ to $10 \times$ in both training and inference.}

Furthermore, \emph{the advantages of the weight-tied model also hold in the multiscale cases} (an improved DEQ variant of~\cite{bai2020multiscale}).
We directly use the available open-source code of \cite{bai2020multiscale} and select two CIFAR models provided therein (i.e.\ MDEQ-Tiny, MDEQ-Large).
The comparison results in \autoref{tab:cifar10_resnet_various_scaling_acc} indicate that the weight-tied model can have more than $3$ times runtime reduction compared to DEQ\footnote{It is worth noting that the hyper-parameters provided in the open-sourced GitHub repository of~\cite{bai2020multiscale} for training these two models are not the same as the one used to retrieve the reported results in the original paper (based on their comments), thus the results may exhibit some differences.}, while maintaining a similar performance compared to the latest DEQ variant (i.e.\ Phantom gradient in~\cite{geng2021training}) and exhibiting a better performance compared to the original DEQ.
Additionally, \emph{Weight-tied model advantage also holds compared to explicit models}.
We select standard ResNet-20 in objective detection and CCT-7 as explicit model baselines and report our result in \autoref{tab:explicit results}.

\begin{table}[!t]
	\caption{\small
		\textbf{On the superior performance of the weight-tied model over other models.}
	}
	\vspace{-0.4em}
	\begin{subtable}[t]{0.49\textwidth}
		\centering
		\caption{\small
			\textbf{Weight-tied model can outperform the same structured DEQ models,} e.g., original DEQ~\cite{bai2020multiscale} and phantom gradient DEQ~\cite{geng2021training} in CIFAR-10 (with data augmentation).
			The weight-tied model has the same structure as the corresponding DEQ and has a depth of $8$.
			In each test, the optimizer and learning hyperparameters for DEQ and the weight-tied model are identical and are chosen based on DEQ's benefit.
			Since the hyperparameters for the original DEQ and phantom DEQ are different, the performance of the original DEQ and phantom gradient DEQ are not comparable.
			\looseness=-1
		}
		\label{tab:cifar10_resnet_various_scaling_acc}
		\resizebox{.8\textwidth}{!}{%
			\begin{tabular}{ccc}
				\toprule
				\textbf{Model Name}                & \textbf{Accuracy} & \textbf{Runtime} \\
				\midrule
				Original Single-stream DEQ         & 81.75\%           & 5.47x            \\
				Single-stream weight-tied          & 84.07\%           & 1x               \\ \midrule
				Original MDEQ Tiny                 & 85.76\%           & 5.03x            \\
				Weight-tied Tiny                   & 85.94\%           & 1x               \\ \midrule
				Original MDEQ Large                & 91.86\%           & 3.4x             \\
				Weight-tied Large                  & 92.36\%           & 1x               \\ \midrule
				Phantom gradient Single-stream DEQ & 85.06\%           & 6.95x            \\
				Single-stream weight-tied          & 86.38\%           & 1x               \\ \midrule
				Phantom gradient MDEQ Tiny         & 88.67\%           & 4.90x            \\
				Weight-tied Tiny                   & 88.59\%           & 1x               \\ \midrule
				Phantom gradient MDEQ Large        & 94.70\%           & 2.69x            \\
				Weight-tied Large                  & 94.54\%           & 1x               \\
				\bottomrule
			\end{tabular}%
		}
	\end{subtable}
	\hfill
	\begin{subtable}[t]{0.49\textwidth}
		\centering
		\caption{\small
			\textbf{Weight-tied model can outperform the similar structured explicit models.}
			This figure shows a performance comparison between the weight-tied model and the explicit model in CIFAR-10 (with data augmentation).
			The notion of \emph{small} or \emph{large} indicates the network width.
			The weight-tied model is a ResNet-like model based on the Single-stream DEQ structure and has the same depth as explicit models.
			The explicit models are dense while the weight-tied model is sparse in order to fit similar performance as explicit models.
		}
		\label{tab:explicit results}
		\vspace{1.em}
		\resizebox{0.95\textwidth}{!}{%
			\begin{tabular}{ccc}
				\toprule
				\textbf{Model Name}                   & \textbf{Accuracy} & \textbf{\# of Param} \\
				\midrule
				ResNet-20 Small                       & 84.07\%           & 68k                  \\
				ResNet Small (weight-tied)            & 84.08\%           & 35k                  \\ \midrule
				ResNet-20 Large                       & 90.96\%           & 4.3M                 \\
				ResNet Large (multi-mask weight-tied) & 90.99\%           & 0.53M                \\ \midrule
				CCT-7 Small                           & 89.43\%           & 0.96M                \\
				CCT Small (weight-tied)               & 89.60\%           & 0.34M                \\ \midrule
				CCT-7 Large                           & 90.11\%           & 3.7M                 \\
				CCT Large (multi-mask weight-tied)    & 90.27\%           & 0.60M                \\
				\bottomrule
			\end{tabular}%
		}
	\end{subtable}
\end{table}

\paragraph{Summary.}
In contrast to the existing line of work like~\cite{bai2019deep,bai2020multiscale,bai2021stabilizing,geng2021training,agarwala2022deep} that aims to enhance the optimization quality of DEQ variants, this paper instead revisits and re-examines their fundamental building block---which is usually overlooked in their investigations---the idea of the weight-tied model.
Surprisingly, as identified in \autoref{fig:cifar10_various_nn_archs_acc_vs_runtime}, \autoref{tab:cifar10_resnet_various_scaling_acc} and \autoref{tab:explicit results}, these weight-tied models are simple yet very effective: \emph{the weight-tied model could outperform most of the latest DEQ variants as well as explicit models in both performance and time complexity across various neural architectures.}
We believe it is worthwhile to leverage the original weight-tied model---which is clean and still an (our newly identified) very strong baseline in this field---to explore other design spaces of implicit models.
\looseness=-1

\section{Multi-Mask Weight-tied Model}  \label{sec:motivation_mm_weight_tied}
In this section, we explore the potential design space using the clean yet effective weight-tied model from the aspect of model capacity/model expressive power.

\subsection{Motivation: Limited Model Capability}
\paragraph{Hypothesis.}
Despite the effectiveness of the weight-tied model, due to the coupled model weights across layers, it is natural to hypothesize that
\begin{center}
	\small
	\emph{
		the model capability of a weight-tied model is largely constrained.
	}
\end{center}

We test this hypothesis using the tool described below.
As stated in \emph{Observation \#1} (in this subsection), the feature representations extracted from each $i$-th tied layer of the weight-tied model exhibit a high degree of similarity, aligning with the hypothesis that \emph{the expressive power of the tied layers are limited and cannot capture distinct feature representations as normal not-tied networks.}

\paragraph{Toolbox.}
Inspired by~\cite{nguyen2020wide}, we utilize the linear version of Centered Kernel Alignment (CKA)~\cite{kornblith2019similarity}, as a robust way to measure the \emph{layer-wise feature activation similarities} of every layer pair, and thus reliably identify architecturally corresponding layers.
These layer similarities will be presented in a squared heatmap, where the similarity between the $i$-th and $j$-th layers is represented in the $(i, j)$ and $(j, i)$ positions of the square.
We briefly outline the formulation of linear CKA below:
\begin{small}
	\begin{equation}
		\textstyle
		\text{CKA}(\mK, \mL) = \frac{ \text{HSIC}( \mK, \mL) }{ \sqrt{ \text{HSIC}( \mK, \mK) \text{HSIC}(\mL, \mL) } } \,,
	\end{equation}
\end{small}%
where $\mK := \mX \mX^\top$ and $\mL := \mY \mY^\top$ represent the similarities between a pair of examples according to the representations in $\mX$ or $\mY$.
The CKA empowers the robust quantitation by normalizing the HSIC metric, a.k.a.\ Hilbert-Schmidt Independence Criterion (measuring the similarity of these similarity matrices).
More details can be found in \autoref{appendix:cka} and~\cite{nguyen2020wide}. When applying CKA to the weight-tied model, we display the output in every tied-layer although physically they belong to the same layer.

\begin{SCfigure}[0.5][!t]
	\vspace{0.5em}
	\includegraphics[width=0.31\textwidth]{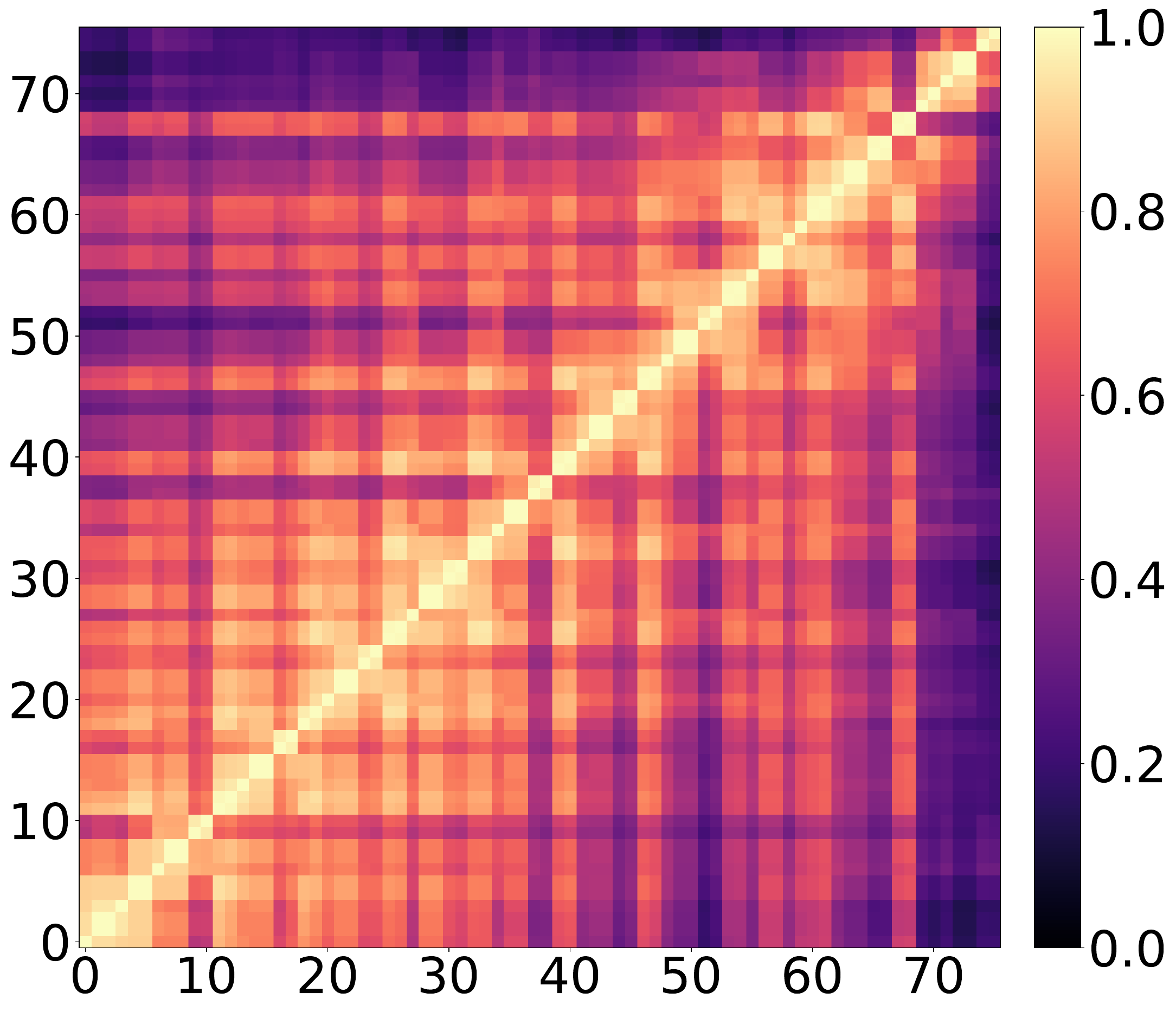}
	\hfill
	\includegraphics[width=0.31\textwidth]{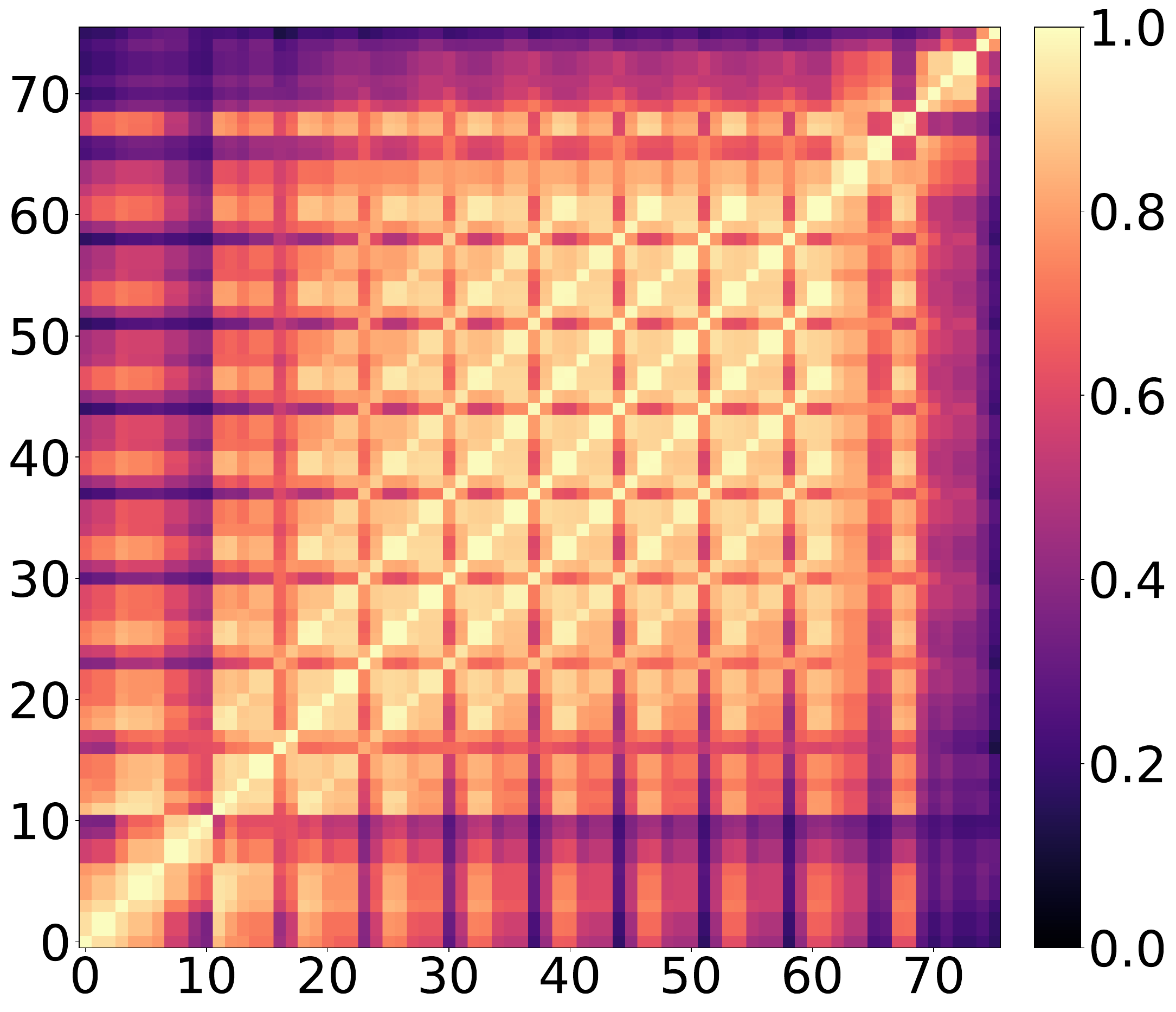}
	\caption{\small
		\textbf{The output similarity pattern} of the same-structured dense normal ResNet model (left) and weight-tied ResNet model (right), with a depth of $8$ trained on the CIFAR-10 classification task.
		The square color in position $(i, j)$ represents the intermediate layer output similarity of the $i$-th layer and $j$-th layer calculated by CKA, and a higher (CKA) value implies higher similarity.
		The weight-tied model has more high-similarity squares.
	}
	\label{cka-weight-tied/not-tied}
\end{SCfigure}

\paragraph{Observation \#1: a high layer-wise similarity emerged in the weight-tied model, indicating a constrained expressive power.}
\autoref{cka-weight-tied/not-tied} depicts a layer similarity heatmap in both the normal (not-tied) model and the weight-tied model.
The patterns shown in the heatmap are distinct between these two models.
Specifically, the weight-tied model illustrates a notable section that displays a \emph{high output similarity}, which can be attributed to the weight-tied module.
This observation suggests that the output produced by the weight-tied module exhibits a higher degree of similarity.
Given the identical parameters in the weight-tied layer, this similarity implies that the weight-tied model is in the way of converging, similar to what happened in DEQ~\cite{bai2019deep}.\looseness=-1

\subsection{Sparsity Trade-off Capacity for Weight-tied Model}
\begin{wrapfigure}{r}{0.4\textwidth}
	\vspace{-1.5em}
	\centering
	\includegraphics[width=0.425\textwidth]{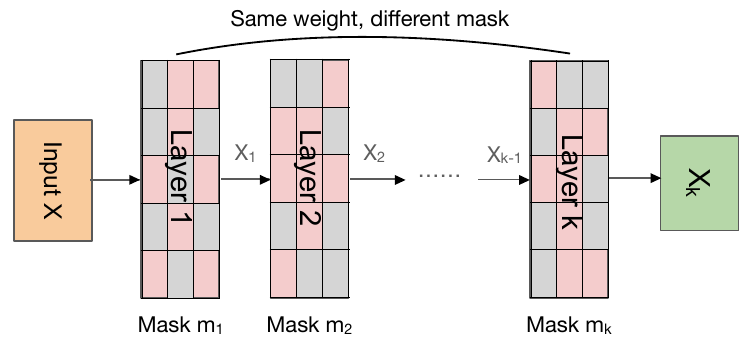}
	\vspace{-.5em}
	\caption{\small
		\textbf{An illustration of multi-mask weight-tied model}.
		The structure shown in the graph is in equivalent form, only one physical layer with $K$ unique and pre-determined boolean masks exists.
	}
	\vspace{-2.05em}
	\label{fig:multi-mask_structure}
\end{wrapfigure}
An intuitive and straightforward idea could be to use diverse sparse masks on the tied layers to induce larger capability for the weight-tied layers, as illustrated in~\autoref{fig:multi-mask_structure}.
We term this design as \emph{\textbf{multi-mask weight-tied}} model, where these boolean masks are distinct, static, and non-trainable across the training, and can be determined before the training phase.
The storage overhead of these boolean masks can be avoided by using \emph{a random generator} to generate deterministic masks on the fly with several scalar seeds per forward and backward pass.

The recursive procedure of a $K$-depth multi-mask weight-tied model can be expressed as,
\begin{small}
	\begin{align}
		\xx_{i + 1} = f_{\ww \odot \mm_{i}}(\xx_{i}) \,,
	\end{align}
\end{small}%
where the masking $\mm_i$ of $i$-th tied layer $\ww$ will only be applied during the forward and backward pass, and the parameter number will reduce with a ratio of $1-s^k$ if masks are generated independently, where $s$ is sparsity ratio of masks.

It is noteworthy to mention the sparsity within the tied layer would significantly trade off the model capability and thus determine the eventual model performance.
We will elaborate on this point in~\autoref{sec:practical_guideline} with detailed practical guidelines.
\looseness=-1

\begin{figure*}[!t]
	\centering
	\vspace{-1.25em}
	\begin{subfigure}{0.325\textwidth}
		\includegraphics[width=1\textwidth]{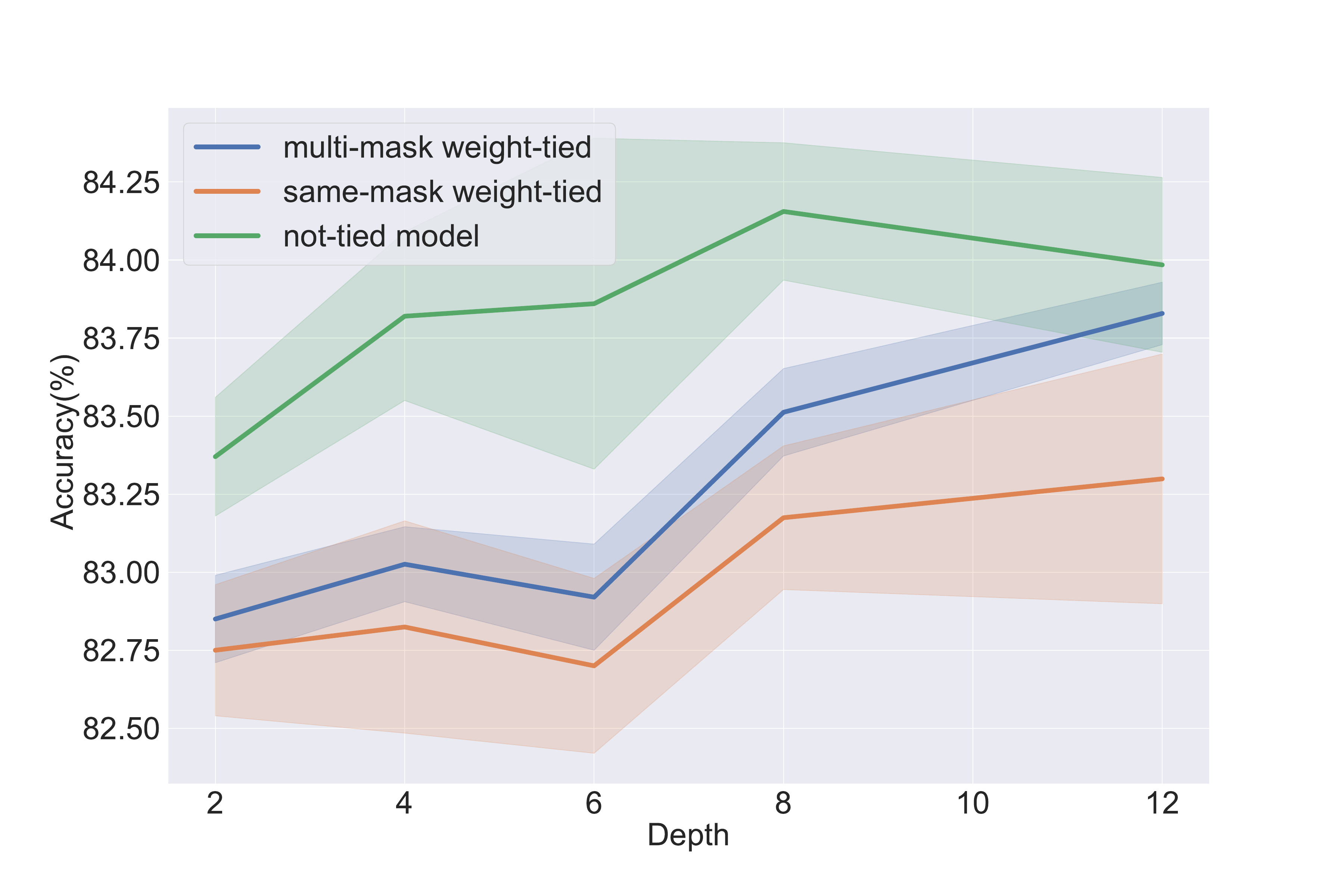}
		\caption{\small Varying depths with density $0.5$.}
	\end{subfigure}
	\hfill
	\begin{subfigure}{0.325\textwidth}
		\includegraphics[width=1\textwidth]{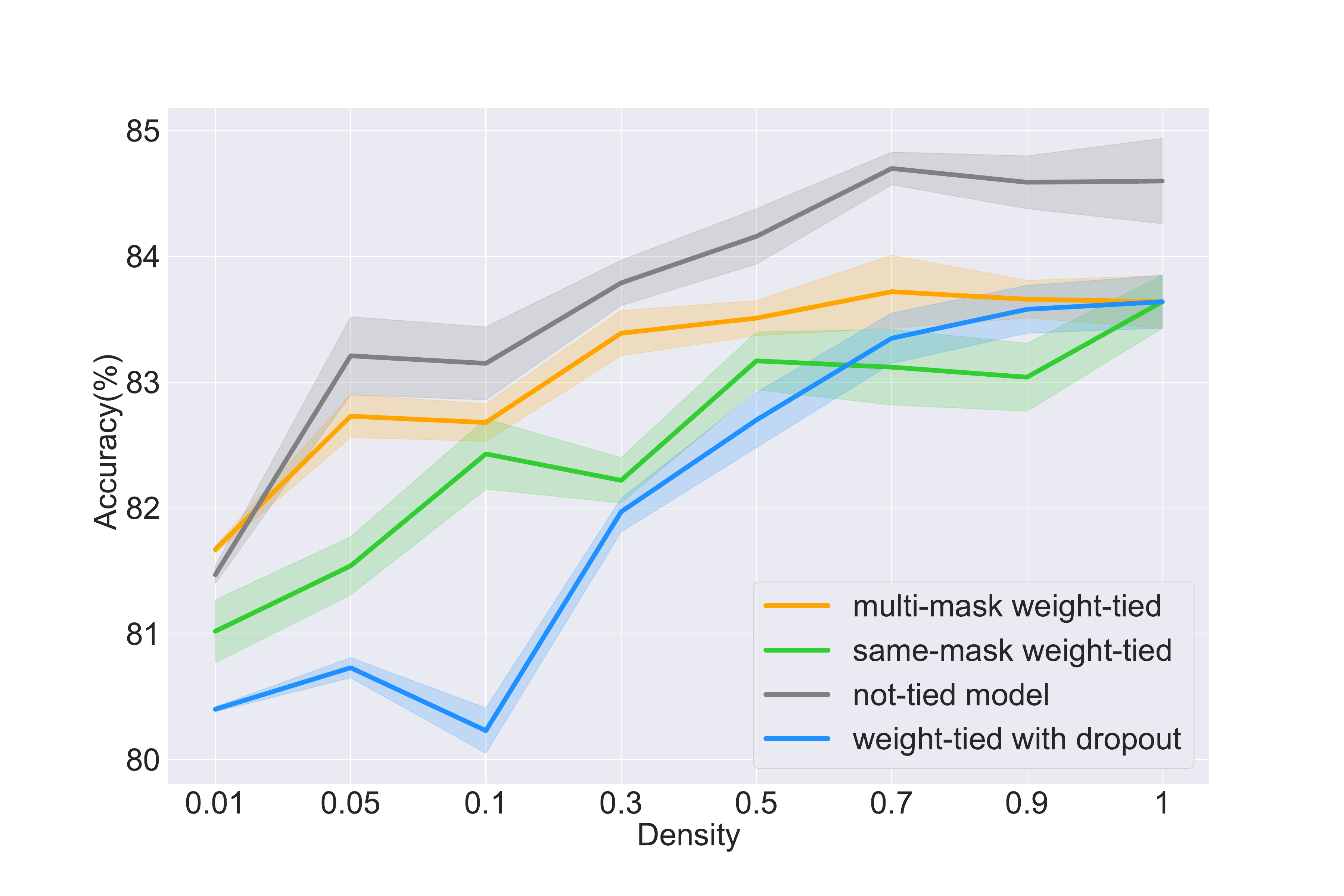}
		\caption{\small Varying densities with depth of $8$}
	\end{subfigure}
	\hfill
	\begin{subfigure}{0.325\textwidth}
		\includegraphics[width=1\textwidth]{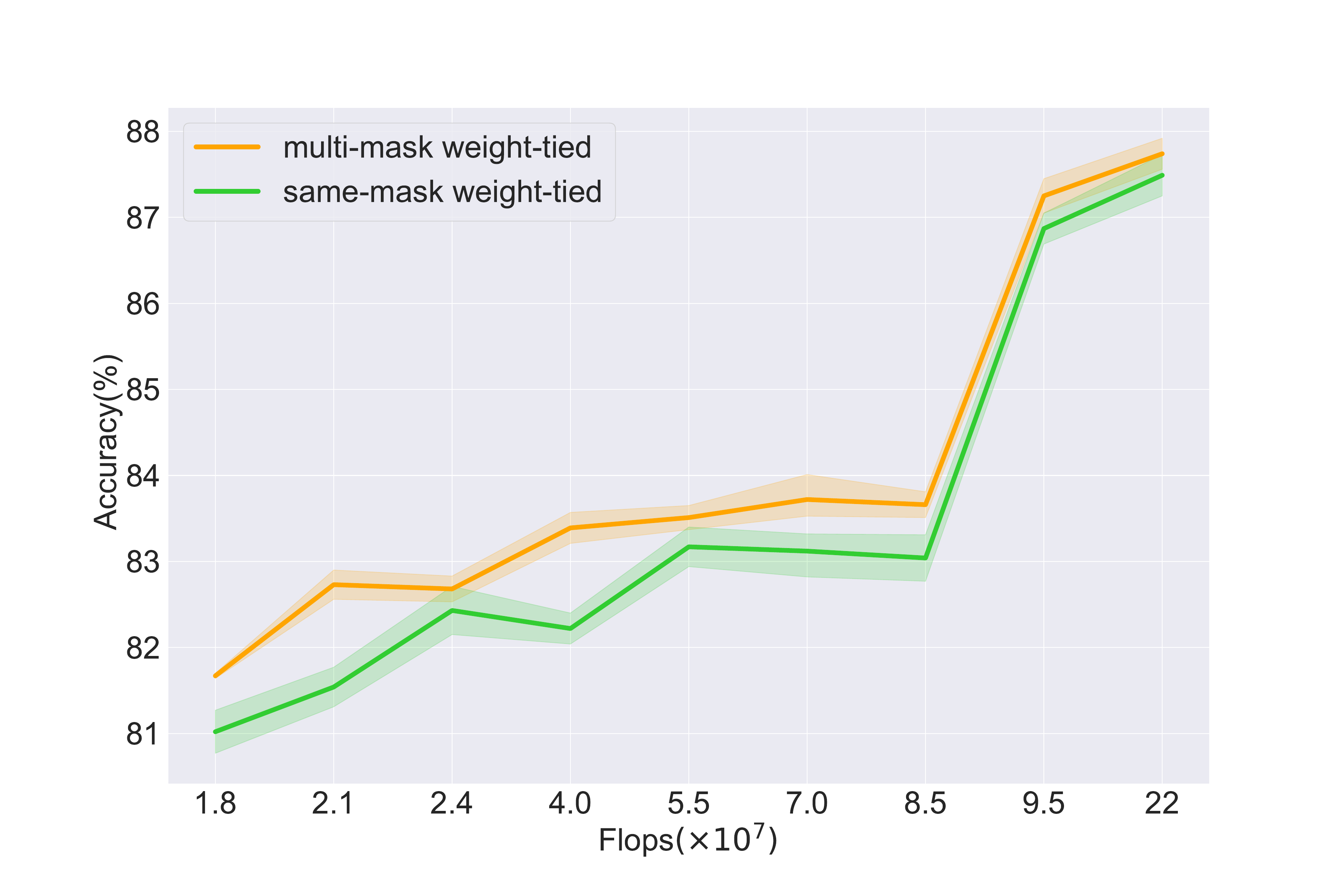}
		\caption{\small Varying FLOPs.}
	\end{subfigure}
	\vspace{-.5em}
	\caption{\small
		\textbf{Multi-mask weight-tied models can surpass same-mask weight-tied models regarding different depths, densities, and FLOPs.}
		All tests are conducted on the CIFAR-10 classification task, for ResNet-like models based on single-stream DEQ structure with 3 independent runs.
		Here the not-tied model has the same structure as the weight-tied model except all layers are not-tied which makes it include d times more parameter numbers.
		Figure (a) is evaluated among different weight-tied depths while all model density=0.5.
		Figure (b) is evaluated among different weight-tied densities while weight-tied depth and corresponding not-tied model depth are equal to 8. Dropout implemented on the weight-tied model is also added as a comparison here.
		Figure (c) is tested among different model widths and weight-tied densities while the weight-tied depth is fixed to 8.
		More details are in the Appendix.
	}
	\vspace{-.5em}
	\label{message2:cifar10_multimask_vs_samemask}
\end{figure*}

\paragraph{Evaluation setup.}
The benefits of the proposed multi-mask weight-tied method can be validated through a fair comparison between (1) the multi-mask weight-tied model, (2) the same-mask weight-tied model, and (3) the conventional model (i.e.\ not-tied model).
Note that here we omit the comparison with DEQ, due to the superior performance of the weight-tied model over DEQ variants as examined in~\autoref{sec:superior_weight_tied_over_deq}.

Following a similar experimental setup in~\autoref{sec:superior_weight_tied_over_deq}, we consider the image classification task on the CIFAR-10 dataset and state some additional treatments for the multi-mask weight-tied model.
For the sake of simplicity, a naive pruning approach referred to as \emph{``equal per layer''}~\cite{frankle2021pruning,price2021dense} was employed to ensure that each layer maintains the same ratio of remaining nodes.
The weight-tied module was the only module that was pruned while other normal modules at the bottom and top layers remain dense.

\paragraph{Observations: \emph{In a variety of depths, densities, and training FLOPs, multi-mask weight-tied models can significantly outperform same-mask weight-tied, and thus justify our intuition.}}
The performance curves regarding various layer depths and layer densities of the weight-tied layers, as well as different numbers of training FLOPs\footnote{
	The sparsity injected in the tied layers would naturally bring an improved efficiency gain.
}, are illustrated in~\autoref{message2:cifar10_multimask_vs_samemask} respectively.
The results show that the multi-mask weight-tied models exhibit a performance benefit of approximately 0.2\% to 1\% in all cases.
In some instances, the multi-mask weight-tied model even surpasses the not-tied model which has 8 times more parameters in the weight-tied layer.
Moreover, regarding the similarity with the dropout method, we also include it in our baseline.
The results also indicate that the multi-mask weight-tied model can also outperform dropout in all cases.

\begin{SCfigure}[2][!t]
	\centering
	\includegraphics[width=0.55\textwidth]{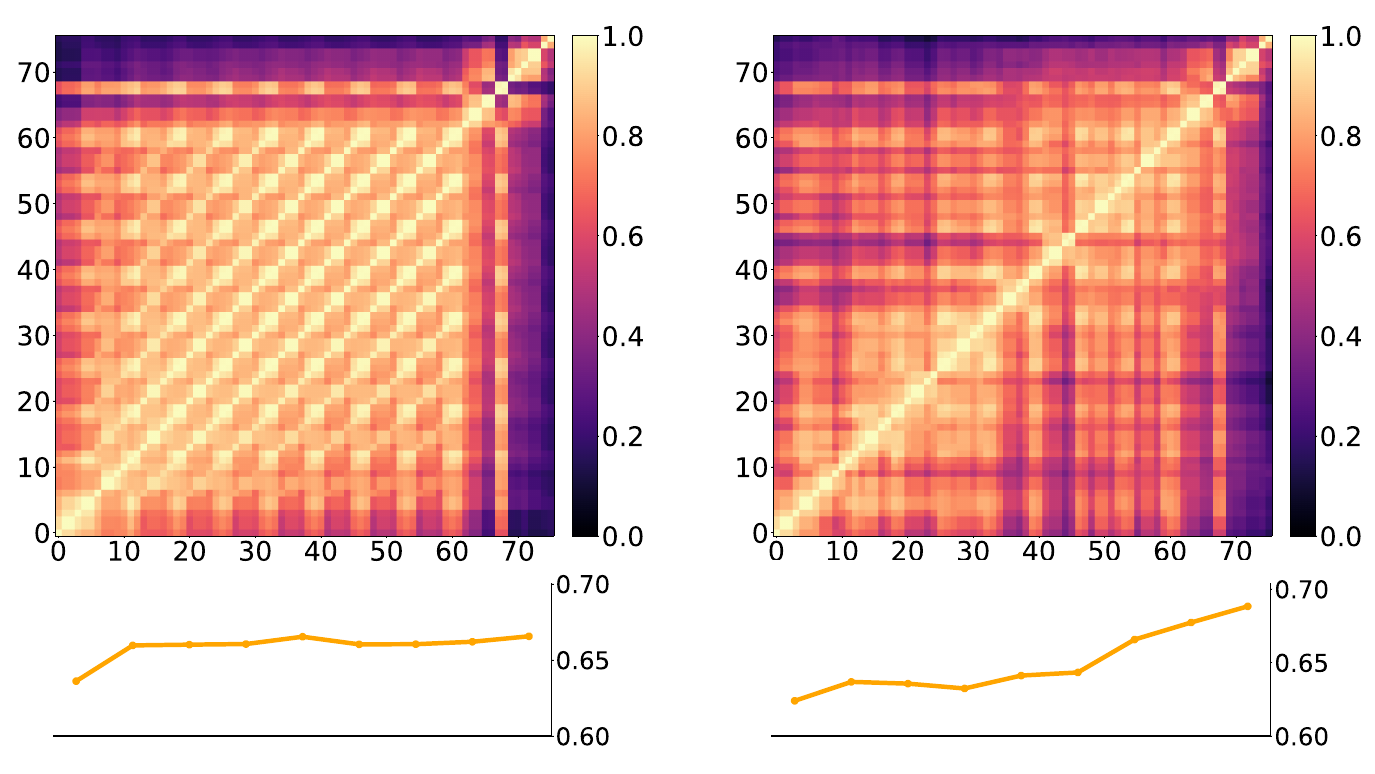}
	\caption{\small
		\textbf{Multi-mask weight-tied approach can erase high output similarity pattern.}
		The figure shows output similarity inside the same-structured same-mask weight-tied model (up left) and multi-mask weight-tied model (upright) on the CIFAR-10 classification task, for models with a density of $0.1$ and depth of $8$.
		The square color in position $(i, j)$ represents layer output similarity in the $i$-th layer and $j$-th layer.
		The bottom figure shows model performance change with layer number increasing.
		The performance is tested through linear probing after each tied layer.
	}
	\label{cka-multi/same_mask}
\end{SCfigure}

\subsection{Interpreting Multi-mask Weight-tied Model}
In previous subsections, motivated by the enormous potential of the weight-tied model in both efficiency and effectiveness, as well as the limited model expressive power constrained by the weight-tied layers, we extend the concept of masking and design a novel multi-mask weight-tied model.
Despite the identified effectiveness and efficiency of the multi-mask weight-tied models, the reasons behind these empirical gains remain unclear.
In this subsection, we further leverage the dynamics in the feature space (using the CKA tool stated in~\autoref{sec:motivation_mm_weight_tied}) to unravel the underlying principles of the multi-mask weight-tied model.

\paragraph{Observation \#2: multi-mask weight-tied approach can erase high output similarity pattern.}
We can witness from \autoref{cka-multi/same_mask} that, \emph{after applying diverse masks to the weight-tied model, the patterns observed in the output similarity squares differ significantly between the case of multi-mask weight-tied model and that of same-mask}.
Specifically, with the multi-mask weight-tied model, the high output similarity region of the weight-tied module disappears, while this high similarity region is accentuated in the case of the same-mask one.

The diminishing of the high output similarity region can be attributed to the diverse sparse mask associated with each tied layer of the multi-mask weight-tied model, resulting in a dissimilar layer in each reused time and thus a decrease in output similarity.
Conversely, when using the same-mask weight-tied model, the sparsified layer remains the same across all weight-tied layers, leading to an increase in output similarity due to the constrained model capability.
Such observation is aligned with our initial intuition of leveraging multi-mask as an implicit way to increase the model's expressive power.
\looseness=-1

The explanation provided above is also consistent with the trend of performance observed in each tied layer illustrated at the bottom of \autoref{cka-multi/same_mask}.
More precisely, we perform Linear Probing used in \cite{kornblith2019similarity} on the trained model per tied layer, as a way to examine the quality of extracted feature representations upon each newly included tied layer.
The performance of the multi-mask weight-tied model continues to improve as the more tied layer is included while that of the same-mask one remains relatively stable: for example, after 8 weight-tied layers, about 3\% performance gain can be observed by the multi-mask one, as compared to the same-mask.

\paragraph{Observation \#3: a larger depth strengthens the high output similarity pattern.}
We can witness from \autoref{cka-depth} of~\autoref{appendix:additional_results} that different weight-tied depths result in a varied level of output similarity in the weight-tied model.
Though models with a higher depth of weight-tied layers exhibit higher similarity, it does not necessarily translate to improved performance, given the degraded performance for depth = 20.
This is likely because the weight-tied layers are close to converging in the early layers and the later layers are unable to make any further contributions to performance, while only increasing the computational costs.

\begin{figure}[!t]
	\centering
	\begin{subfigure}{0.49\textwidth}
		\centering
		\includegraphics[width=0.8\textwidth]{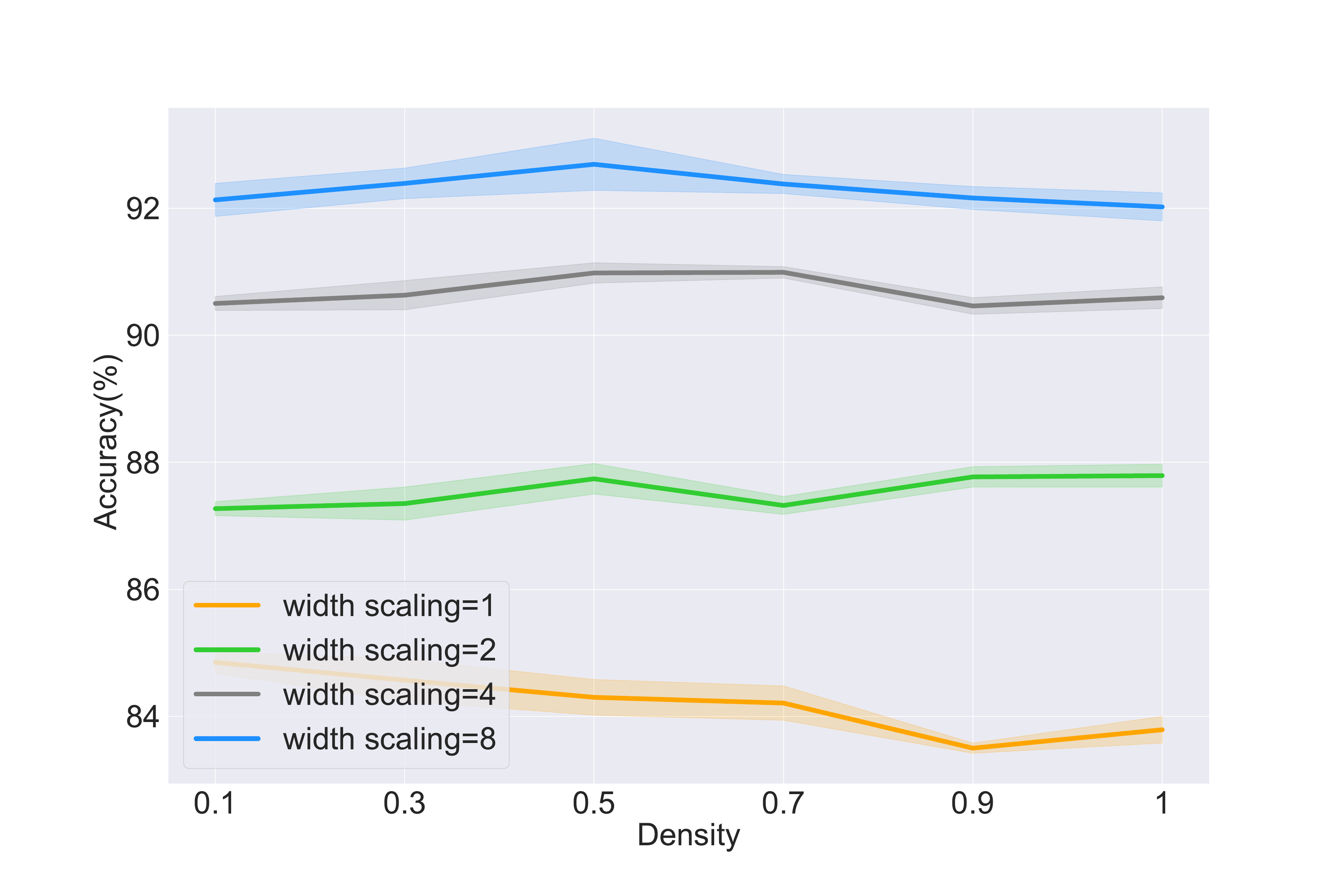}
		\caption{\small ResNet-like model.}
		\label{tradeoff-density}
	\end{subfigure}
	\hfill
	\begin{subfigure}{0.49\textwidth}
		\centering
		\includegraphics[width=0.8\textwidth]{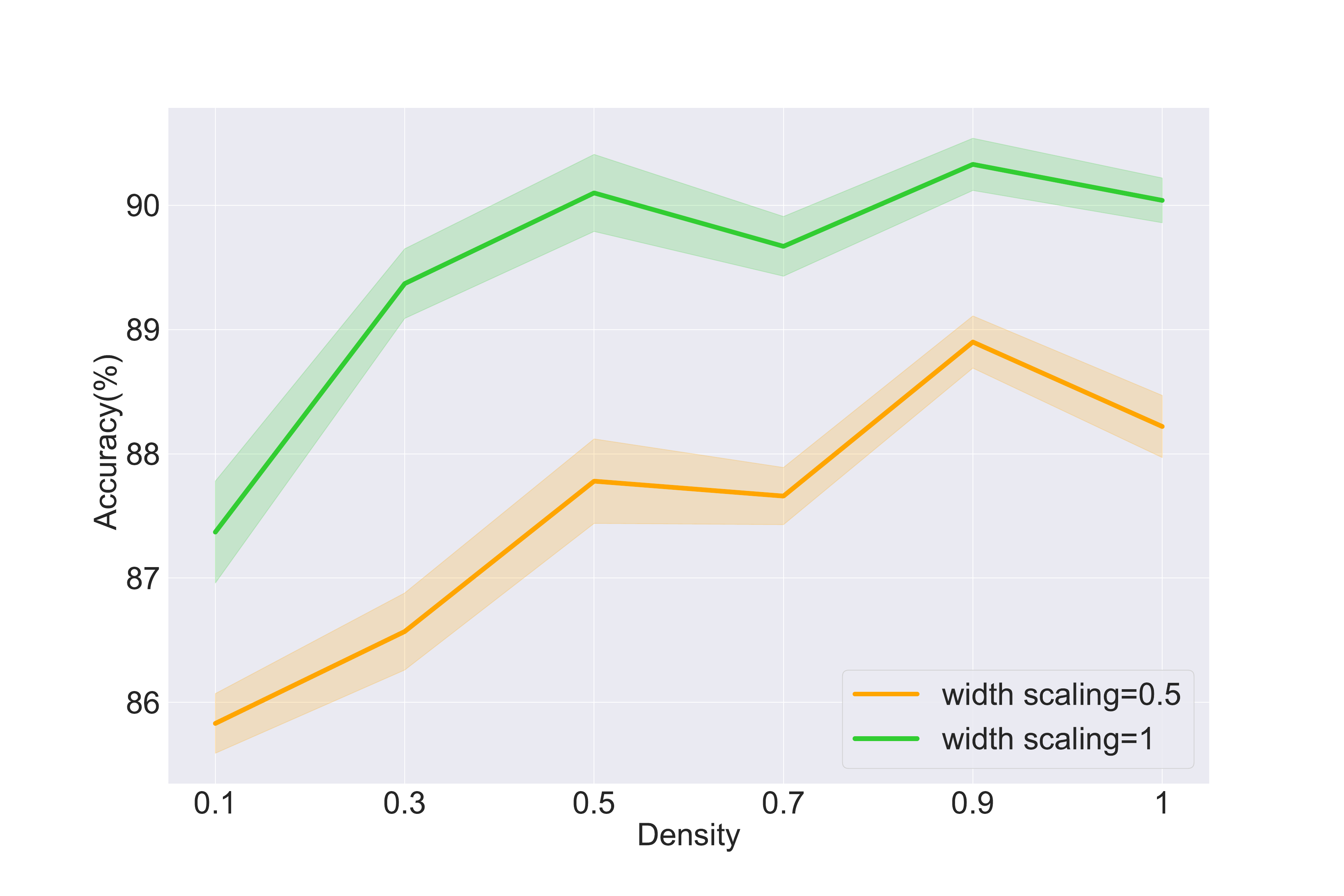}
		\caption{\small CCT-like model.}
		\label{tradeoff-cct}
	\end{subfigure}
	\vspace{-0.5em}
	\caption{\small
		\textbf{Sparse multi-mask weight-tied model can outperform dense weight-tied model.}
		We evaluate model performance under different densities and model widths, with a fixed depth for the CIFAR-10 classification task ($8$ for the ResNet-like model and $7$ for the CCT-like model).
		Models are trained in the same number of FLOPs.
	}
\end{figure}

\begin{SCfigure}[2][!t]
	\centering
	\vspace{-1em}
	\includegraphics[width=0.475\textwidth]{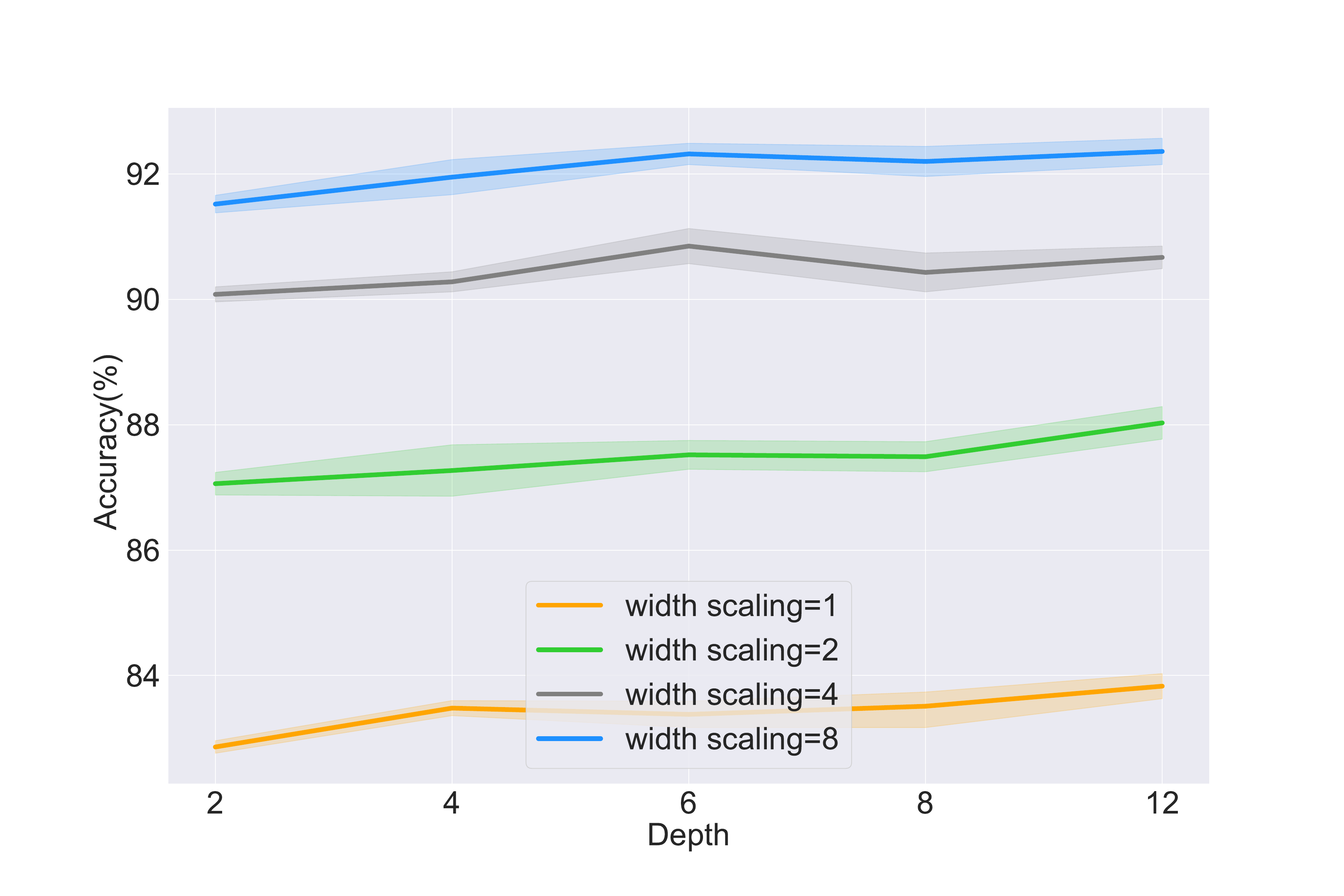}
	\caption{\small
		\textbf{Model width is more essential to performance than weight-tied depth.}
		All tests are conducted in ResNet-like models on the CIFAR-10 classification task.
		We evaluate model performance under different model widths and weight-tied depths while the weight-tied density is fixed at 0.5.
		Model width is controlled by multiplying every layer with a width scaling factor.
		\looseness=-1
	}
	\vspace{-1.em}
	\label{tradeoff-depth}
\end{SCfigure}

\section{Trade-offs and Practical Guidelines} \label{sec:practical_guideline}
Despite the empirical effectiveness of the multi-mask weight-tied model, it may be non-trivial to identify a proper configuration in practice. When transforming a conventional neural architecture to a multi-mask weight-tied model, there exist at least three hyperparameters to configure, namely the depth, mask density, and model width, of the multi-mask weight-tied part.
A different combination of these parameters can result in noticeable variations and trade-offs in model performance.
Therefore, in this section, we provide some crucial insights to study the trade-off of these parameters and provide a guideline for practitioners.

\subsection{Don't Increase Model Depth, Increase Model Width} \label{sec:width_matters}
\paragraph{Rather than the depth of weight-tied layers, model width is more essential to the model performance.}
\autoref{tradeoff-depth} illustrates the performance under various weight-tied depths and model widths (we fix model density to 0.5 to avoid the influence of mask density).
We can witness that \emph{the benefits brought by increasing the depth in the weight-tied layers are far behind that of increasing the model width}, where a 12-depth multi-mask weight-tied model significantly lags behind a 2-depth multi-mask weight-tied which is 2 times wider.
Furthermore, a large value of depth in the weight-tied layer may not always correspond to the improved performance: it intuitively explains DEQ's difficulty in (significantly) surpassing the simple weight-tied model in \autoref{fig:cifar10_various_nn_archs_acc_vs_runtime}, in which an infinite depth may not guarantee better performance than finite depth.

\paragraph{Practical guide \#1:}
The model width matters, rather than the depth of weight-tied layers: practically it is sufficient to use a depth of $6$ or $8$ in the weight-tied layers to ensure reasonably good performance. \looseness=-1

\begin{figure}[!t]
	\centering
	\begin{subfigure}{0.475\textwidth}
		\centering
		\includegraphics[width=0.65\textwidth]{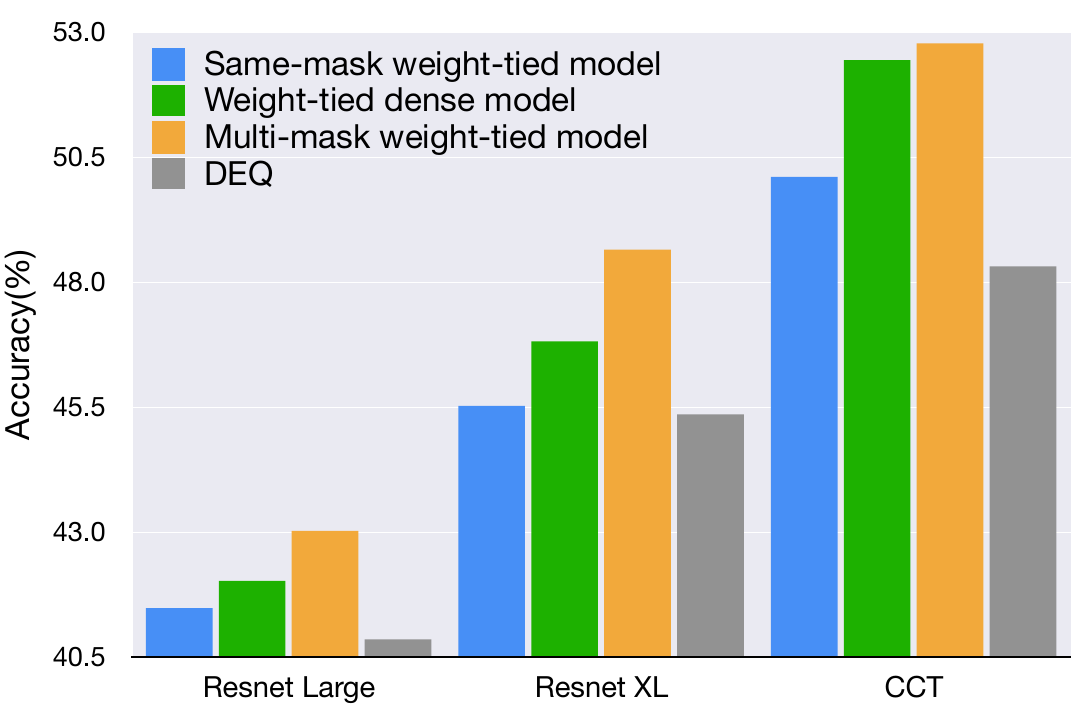}
		\caption{\small
			Multi-mask weight-tied model surpasses the dense weight-tied model, same-mask weight-tied model, and DEQ, for ResNet-like and CCT-like architecture on ImageNet32.
			Weight-tied models use a density of 0.5 and a depth of 8.
		}
		\label{imagenet}
	\end{subfigure}
	\hfill
	\begin{subfigure}{0.5\textwidth}
		\centering
		\resizebox{1.\textwidth}{!}{%
			\begin{tabular}{ccc}
				\toprule
				\textbf{Mask density} & \textbf{CIFAR-10 Acc (\%)} & \textbf{ImageNet32 Acc (\%)} \\
				\midrule
				0.3                   & $90.63$                    & $42.97$                      \\
				0.5                   & $\mathbf{90.98}$           & $\mathbf{43.05}$             \\
				0.7                   & $\mathbf{90.99}$           & $42.94$                      \\
				0.9                   & $90.46$                    & $42.66$                      \\ \midrule
				1                     & $90.59$                    & $42.04$                      \\
				\bottomrule
			\end{tabular}%
		}
		\vspace{1.5em}
		\caption{\small
			Multi-mask weight-tied model surpasses the dense weight-tied model, for ResNet-Large models under different densities with a fixed depth of $8$.
			\looseness=-1
		}
		\label{tab:imagenet}
	\end{subfigure}
	\vspace{-0.5em}
	\caption{\small
		\textbf{Sparse multi-mask weight-tied model can outperform other variants on ImageNet32.}
		ImageNet32 is a down-sampled ImageNet dataset including all ImageNet images.
		ResNet-XL is two times wider than ResNet-Large.
		All models are trained in the same number of FLOPs.
	}
\end{figure}

\subsection{On the Superior Empirical Effectiveness of Multi-mask Weight-tied Model}
Given the practical guide \#1 found in~\autoref{sec:width_matters}, in this section, we consider experiments for the multi-mask weight model with a fixed weight-tied depth while varying the mask densities and model widths. \looseness=-1

\paragraph{Sparse multi-mask weight-tied model can outperform dense weight-tied model.}
In addition to the effectiveness of the multi-mask weight-tied model identified in~\autoref{message2:cifar10_multimask_vs_samemask}, the results in~\autoref{tradeoff-density} move one step further by comparing the sparse multi-mask weight-tied model with the dense counterparts.
The superior effectiveness of the multi-mask weight-tied model can be justified by the fact that:
\emph{when trained with the same number of FLOPs, the multi-mask weight-tied model can outperform the dense weight-tied model in each model width} (similar pattern persists when training models with the same amount of training epochs).
Such an observation is also consistent across different neural architectures, as verified by a Vision-Transformer model in~\autoref{tradeoff-cct}.
\looseness=-1

\paragraph{On the hardware-friendly multi-mask weight-tied model.}
The 2:4 fine-grained structured sparse network~\cite{choquette2021nvidia,zhang2022learning} trades off the benefits of both unstructured fine-grained sparsity and structured coarse-grained sparsity, by accelerating matrix multiplication at least two times through NVIDIA's sparse tensor cores~\cite{pool2021accelerating}.
As the 2:4 fine-grained structured sparsity inherently exhibits 50\% sparsity, which aligns with our suggested sparsity ratio for the multi-mask weight-tied model,~\autoref{tab:NM sparsity} demonstrates that \emph{our multi-mask weight-tied model unleash the potential of delivering performance gains across sparsity structures, while maintaining high levels of computation efficiency and learning effectiveness}.

\begin{table}[!t]
	\small
	\caption{\small
		\textbf{Multi-mask remains superior performance with 2:4 fine-grained structured sparsity} on CIFAR-10 classification task.
		The ``Small'' and ``Large'' refer to different model widths.
		Results are averaged over three trials. \looseness=-1
	}
	\vspace{-0.1em}
	\label{tab:NM sparsity}
	\centering
	\resizebox{0.65\textwidth}{!}{%
		\begin{tabular}{ccccc}
			\toprule
			\multirow{2}{*}{ \textbf{Model Size} } & \multicolumn{2}{c}{ \textbf{ResNet-like model} } & \multicolumn{2}{c}{ \textbf{CCT-like model} }                                                                                       \\ \cmidrule(lr){2-3} \cmidrule(lr){4-5}
			                                       & 2:4 sparsity                                     & unstructured sparsity                         & 2:4 sparsity                             & unstructured sparsity                    \\
			\midrule
			Small                                  & $83.16_{ \transparent{0.5} \pm 0.10 }$\%         & $83.51_{ \transparent{0.5} \pm 0.14 }$\%      & $88.08_{ \transparent{0.5} \pm 0.63 }$\% & $87.78_{ \transparent{0.5} \pm 0.34 }$\% \\
			\midrule
			Large                                  & $90.44_{ \transparent{0.5} \pm  0.21 }$\%        & $90.43_{ \transparent{0.5} \pm 0.31 }$\%      & $89.56_{ \transparent{0.5} \pm 0.46 }$\% & $90.10_{ \transparent{0.5} \pm 0.31 }$\% \\
			\bottomrule
		\end{tabular}%
	}
\end{table}

\paragraph{Practical guide \#2:}
Instead of using a dense weight-tied model, the multi-mask weight-tied model is a more appealing choice.
A density of $0.5$ usually achieves an optimal performance across all densities for ResNet-like architectures, while a density of $0.9$ might be more suitable for Transformer-like architectures.

\subsection{Examining the Generalizability of the Findings on ImageNet}
In this subsection, we examine the effectiveness and generalizability of the two key findings for the multi-mask weight-tied model, on the challenging ImageNet\footnote{
	Due to the computational feasibility, we only afford to evaluate on a down-sampled ImageNet.
	However, we believe the success therein can be transferred to the original ImageNet as well as other large-scale datasets.
} dataset for both ResNet-like and Transformer-like neural architectures.
In \autoref{imagenet}, the multi-mask weight-tied model again outperforms both the dense weight-tied model and the sparse weight-tied model with the same mask across layers.
In \autoref{tab:imagenet}, the sparse multi-mask weight-tied model can still outperform the dense weight-tied model in ImageNet32 and reach the best performance at the density of 0.5.

\section{Discussion and Conclusion}
In this paper, we first identify that DEQ is subjected to model inefficiency and optimization instability. To address these limitations, we revisit implicit models and trace them back to the original weight-tied models. Our experiments indicate that weight-tied models can outperform existing DEQs in terms of both performance and computational expense. To further enhance the model capacity of weight-tied models, we propose the use of multi-mask weight-tied models. The superior performance and effectiveness of multi-mask weight-tied models, in comparison to same-mask weight-tied models, dense weight-tied models, and DEQs, are established by empirical experiments across various model structures and tasks. To facilitate the practical use of multi-mask weight-tied models, we examine the trade-off between depth, width, and sparsity of the weight-tied layer and indicate that relatively wider and sparser models are preferred.
\looseness=-1

\clearpage
\bibliography{paper}
\bibliographystyle{abbrv}

\clearpage
\appendix

\begingroup
\setlength{\parskip}{4pt plus4pt minus0pt}

\section{Related Work} \label{appendix:complete_related_work}
\paragraph{Implicit models and DEQ variants.}
As a way of replacing explicit layers with one implicit layer and prescribed internal dynamics, implicit models have attracted wide attention from the community in recent years~\cite{amos2017optnet,chen2018neural,niculae2018sparsemap,wang2019satnet,bai2019deep,bai2020multiscale,bai2021stabilizing,geng2021training,agarwala2022deep}.
The DEQ model, as introduced by~\cite{bai2019deep}, is one representative approach in implicit models that finds the equilibrium of a system to eventually reach a fixed point equation.
Rather than considering the orthogonal, computationally expensive, or numerical unstable Neural ODE~\cite{chen2018neural}, in this work, we take the more scalable and promising line of the deep equilibrium approach as stated in the literature~\cite{bai2020multiscale,geng2021training,pokle2022deep}, to examine the latest progress in implicit models.

DEQ, in it's original form~\cite{bai2019deep}, severely suffers from the issues like training instability and computational inefficiency.
The perspective of training instability was later discussed in~\cite{bai2021stabilizing} by proposing a regularization scheme for the ill-conditioned Jacobian to stabilize the learning.
The work of~\cite{agarwala2022deep} theoretically discusses this instability from the view of initialization statistics, though only toy examples are provided on MNIST with no significant gains being observed when compared to the best-performing networks.
The work of~\cite{geng2021training} instead pursues to approximate the exact calculation of the Jacobian-inverse term and thus accelerates the training by at most $1.7~\times$ while still achieving performance on par with that of DEQ variants.
Regarding deep learning applications, the work of~\cite{bai2020multiscale} propose \emph{Multiscale DEQ} (MDEQ) to improve upon image classification, while~\cite{pokle2022deep} take all recent ingredients of DEQ variants and adapt DEQ to diffusion models.

However, it is noteworthy to mention that these recent researches largely ignore the original weight-tied model, despite being simple, effective, and memory inefficient, making the generalizability and practicality of DEQ variants on various use cases to be questioned; our contribution therein.

\paragraph{Weight-tied model.}
Weight-tied models, often referred to as weight-sharing models, are a popular paradigm to achieve parameter-efficient feature. These models employ a unified set of weights across diverse layers to largely reduce parameter numbers~\cite{dehghani2018universal,dabre2019recurrent, xia2019tied, lan2020ALBERT,li2021training,takase2021lessons}.
Serving as the key of numerous implicit models, they have been subject to extensive investigation in recent years across a range of applications~\cite{wang2019weight,liu2020comprehensive,yang2018unsupervised,lan2020ALBERT,takase2021lessons,zhang2020deeper,bender2020can,xie2021weight,li2021training}.
For example, the Universal Transformer, initially introduced in~\cite{dehghani2018universal} innovatively ties all parameters within a single Transformer layer.  This notion is subsequently adopted in related studies such as~\cite{dabre2019recurrent,lan2020ALBERT}.
Furthermore, the work of~\cite{xia2019tied} further extends parameters sharing between the encoder and decoder components in encoder-decoder structure.
In a different vein, ~\cite{xiao2019sharing} proposes a attention weights tying strategy to enhance the computational efficiency of Transformers.
Building on these ideas, ~\cite{takase2021lessons} takes parameter tying to the next level by suggesting three strategies for tying the parameters of various layers with various combinations, transcending the approach of merely sharing parameters from one layer across all layers. \looseness=-1

While the existing researches for the weight-tied model primarily concerns methods for tying diverse layers, they do not encompass the introduction of sparse prunning masks to a shared layer, as in our approach.
Separately, in the context of Neural Architecture Search (NAS), weight-sharing methods are applied to samples distinct neural architectures from a super net with sparse masks to alleviate computational burdens. In this setup, an abundance of architectures can share weights within the same super net and the expensive training procedure can also be reduced to only once.~\cite{zhang2020deeper,bender2020can,xie2021weight}

\paragraph{Model quantization and pruning.}
A line of seminal papers for model quantization~\cite{han2015deep,chen2015compressing} employs the concept of hash functions or quantization to map weights to scalars or codebooks, thereby increasing the compression rate. This approach has been further extended to soft weight sharing~\cite{li2020group,ye2018unified,ullrich2017soft,zhang2018learning}, where the remaining weights are assigned to the most probable clusters. However, this strategy differs from our approach of using sparse masks to enhance capability.

In the realm of model pruning~\cite{wang2023state}.
three main avenues have emerged: i) pruning when initializing, ii) dynamic pruning during training, and iii) pruning after training. The latter two typically involve pruning model weights with extra training or calculation and thus not efficient.
The method of pruning when initializing first replies on magnitude-based metrics to do pruning~\cite{frankle2018the}. However, several subsequent studies~\cite{su2020sanity,frankle2021pruning,wang2022recent} have ignited a debate, asserting that random masks - randomly sample pruning mask without any prior knowledge - can be just as effective as the earlier "lottery ticket" idea~\cite{frankle2018the}. 
It's worth noting that most model pruning researches traditionally focus on improving conventional explicit neural networks by refining pruning criteria and proposing advanced optimization strategies or objective functions. As far as our knowledge extends, the introduction of various random masks into a weight-tied model, as presented in our work, is a novel concept.

A related work that bears relevance to our manuscript is ~\cite{bai2022parameter}, which has only one physical layer and employs masks atop this fixed layer to generate diverse dense layers.
In particular, this approach utilizes several unique masks to select different sets of values from a random vector (i.e., codebook), thereby creating distinct dense layers.
Nevertheless, it remains distinct from our fundamental idea of utilizing weight-tied structure to learn a tied weight with deterministic random binary masks, which implicitly imparts model capacity. \looseness=-1

\paragraph{Drouput.}
Dropout, introduced by~\cite{hinton2012improving} serves as a pivotal training technique aimed at mitigating overfitting~\cite{labach2019survey,liu2023dropout}. It achieves this by introducing random modifications to neural network parameters or activations~\cite{wan2013regularization,ba2013adaptive,wang2013fast,kingma2015variational,gal2016dropout}.
While Dropout has found application in compressing neural networks~\cite{molchanov2017variational,neklyudov2017structured,gomez2019learning}, it's important to note that the \emph{stochastic} dropping idea in Dropout is primarily tailored for standard, explicit neural architectures, which stands in contrast to the deterministic masks of our weight-tied models.

\section{Toolbox: CKA} \label{appendix:cka}
Centered Kernel Alignment, proposed by \cite{nguyen2020wide}, is a representation similarity measurement. It is invariant towards linear transformation, orthogonal transformation as well as isotropic scaling. We briefly outline the formulation of linear CKA below, where the CKA empowers the robust quantitation by normalizing the Hilbert-Schmidt Independence Criterion (HSIC) metric:
\begin{small}
	\begin{equation}
		\textstyle
		\text{CKA}(\mK, \mL) = \frac{ \text{HSIC}( \mK, \mL) }{ \sqrt{ \text{HSIC}( \mK, \mK) \text{HSIC}(\mL, \mL) } } \,.
	\end{equation}
\end{small}%
Note that $\mX \in \R^{m \times p_1}$ and $\mY \in \R^{m \times p_2}$ contain representations of two layers, with $p_1$ and $p_2$ neurons respectively. Each element of the $m \times m$ Gram matrices $\mK := \mX \mX^\top \in \R^{m \times m}$ and $\mL := \mY \mY^\top \in \R^{m \times m}$ represents the similarities between a pair of examples according to the representations in $\mX$ or $\mY$. \\

HSIC measures the similarity of centered similarity matrices and thus is invariant to orthogonal transformations of the representations as well as to permutation of neurons, namely $\text{HSIC}(\mK, \mL) := \frac{1}{{(m-1)}^2 } \text{vec}(\mH \mK \mH) \text{vec}(\mH \mL \mH)$, where $\mH = \mI - \frac{1}{n} \1 \1^\top$ is a centering matrix. More details can be found in~\cite{nguyen2020wide}.

\section{Experiment Details} \label{appendix:exp_details}
\subsection{Experiment of Fig. \ref{fig:cifar10_various_nn_archs_acc_vs_runtime}}
For our baseline experiments, we use the open-source implementations provided by the respective authors.
For MDEQ \cite{bai2020multiscale}, we follow the settings provided in its open-source code and train the model using Adam optimizer with the cosine learning rate scheduler.
The maximum learning rate is set to $0.001$ with a weight decay of $2.5$e$-6$. The batch size is 64. 
During experimentation with the CIFAR-10 dataset, augmentations in the form of basic normalization, random cropping, and horizontal flipping are applied.

For different type of models, the settings are slightly different.
\begin{itemize}[leftmargin=12pt]
	\item For MLP models in CIFAR-10, we use Adam optimizer with a cosine learning rate scheduler.
	      The maximum learning rate is 0.001 with a weight decay of $2.5$e$-6$.
	\item For ResNet models in CIFAR-10, we use Adam optimizer with a cosine learning rate scheduler.
	      The maximum learning rate is 0.001 with a weight decay of $2.5$e$-6$.
	      The model is trained for 150 epochs and batch size is 128.
	\item For CCT models in CIFAR-10, the AdamW optimizer is used with a cosine learning rate scheduler.
	      The maximum learning rate is 6e-4 and the model is trained for a total of 300 epochs, following CCT open-source code default settings.
\end{itemize}

In all the experiments, we keep the DEQ model and weight-tied model in same structure and same parameter number. The only difference is that we use weight-tied module to replace previous DEQ module in DEQ models. 
The weight-tied module and DEQ module also have the same model structure, but utilize different input method. DEQ use $\zz$ as input and inject $\xx$ into every layer, while weight-tied module only use $\xx$ as input and discard $\zz$.
We will refer this kind of weight-tied model as weight-tied version DEQ model in the following paragraphs.

\subsection{Experiment of Table. \ref{tab:cifar10_resnet_various_scaling_acc}}
The implementations of original DEQ and DEQ with~\textit{phantom gradient} are based on open source provided by respective authors.
For MDEQ \cite{bai2020multiscale}, we train the model using Adam optimizer with the cosine learning rate scheduler.
The maximum learning rate is set to $0.001$ with a weight decay of $2.5$e$-6$. The batch size is 64. We test all three provided DEQ models(single-stream, Tiny and Large) and corresponding weight-tied verson models.
Same size DEQ model and weight-tied model are in same structure and same parameter number. They also use same optimizer and hyperparameters. The model details can be find in \cite{bai2020multiscale}.

For DEQ with~\textit{phantom gradient}, we follow the setting provided in \cite{geng2021training} and use SGD with the cosine learning rate scheduler.
The maximum learning rate is set to 0.2 with a weight decay of 0.0001. We also test Single-stream, Tiny and Large three different size of models.

For both the MDEQ and DEQ with~\textit{phantom gradient}, the total epoch number is set to 50, 50, and 220 for Single-stream, Tiny and Large DEQ respectively. 

\subsection{Experiment of Table. \ref{tab:explicit results}}
In ResNet comparison, we select standard ResNet-20 in objective detection as baseline explicit model and single-stream DEQ structure as weight-tied model structure. 
ResNet-20 includes 4 BasicBlocks in total, and the DEQ module in Single-stream DEQ is also a BasicBlock. To make fair comparison, we make this DEQ module as 4-depth weight-tied module, which reuses this Basicblock 4 times.
Because of this reusing, weight-tied models enjoy much less parameter numbers.

In CCT comparison, both explicit model and weight-tied model use CCT-7 model structure \cite{hassani2021escaping}. The weight-tied models also have depth of 4.

\subsection{Experiment of Fig. \ref{message2:cifar10_multimask_vs_samemask}}
All tests in Fig. \ref{message2:cifar10_multimask_vs_samemask} are conducted on the CIFAR-10 classification task with 3 independent runs. The base model is weight-tied version Single-stream DEQ model.
Multi-mask weight-tied model utilizes different pruning masks in different tied layers, while same-mask weight-tied model utilizes only one pruning mask in all tied layers. Not-tied model is an explicit model which looses the tie of different tied-layers.
The parameter numbers of not-tied model is d times larger compared to weight-tied model. In training, we use Adam optimizer with a cosine learning rate scheduler. The training epoch is 150.

Figure (a) varies the depth of weight-ied model and Figure (b) varies the mask density of weight-ied model. Based on the data in Figure(a) and Figure(b), we create Figure(c) whose x-asis is FLOPs. 

\subsection{Experiment of Fig. \ref{tradeoff-density} and Fig. \ref{tradeoff-depth}}
In Fig. \ref{tradeoff-density}, model performance are evaluated under different densities and model widths, with a fixed depth for the CIFAR-10 classification task.
The weight-tied depth of ResNet model is 8 and depth for CCT model is 7. The base models are weight-tied version Single-stream DEQ model and CCT-7 respectively. Models are trained in the same number of FLOPs

In Fig. \ref{tradeoff-depth}, model performance are evaluated under different weight-tied depth and model widths, with a fixed model density of 0.5.

\subsection{Experiment of Fig. \ref{imagenet}}
For experiment in imagenet, the experiment settings are listed as follows. 
\begin{itemize}[leftmargin=12pt]
	\item For Resnet models in ImageNet32, we use the SGD optimizer with a multi-step learning rate scheduler.
	The maximum learning rate is 0.05 with a weight decay of 1e-4.
	The total epoch number is 90 and the batch size is 128.
	\item For CCT models in ImageNet32, we again use AdamW with a cosine learning rate scheduler.
	The maximum learning rate is 5e-4 with a total training epoch of 300 instead.
\end{itemize}
Different methods in same model type are in same model structure and same parameter numbers. They are also with same optimizer and hyperparameters. Weight-tied model depth is fixed with 8.

\section{Additional Results} \label{appendix:additional_results}
\subsection{On the Ineffectiveness of DEQ} \label{appendix:ineffectiveness_DEQ}

\paragraph{Computational inefficiency.}
Despite the elegance of the DEQ concept, the gradient estimation of these implicit models bottlenecks their practicality, due to the deficiency in both training and inference efficiency caused by the expensive Jacobian-inverse term or iterative Jacobian-vector products.
For example, the Broyden solver used in DEQ~\cite{bai2019deep} for exact gradient estimation would usually introduce over $30$ iterations in the backward pass, leading to a prohibitive cost and causing severe slow-down when compared with standard explicit models.
Though the implicit gradient estimation method (a.k.a.\ phantom gradient) proposed in \cite{geng2021training} fastens~\cite{bai2019deep} by at most $1.7 \times $, it does not fully mitigate the efficiency issue due to the noticeable efficiency gap in~\autoref{fig:cifar10_various_nn_archs_acc_vs_runtime} (even after $1.7 \times$ acceleration).

\paragraph{Optimization instability.}
In addition to the computational inefficiency, the training process of DEQs also exhibits instability~\cite{agarwala2022deep}.
Many DEQ variants, such as the original DEQ~\cite{bai2019deep,bai2020multiscale} or DEQ with Jacobian regularization~\cite{bai2021stabilizing}, indeed require employing a pre-training step by reusing DEQ layers in a weight-tied manner (stated in~\autoref{sec:intro_to_weight_tied}), before entering the formal DEQ training with root-finding solvers.
Such transition normally results in a remarkable performance drop, e.g.\ an approximated 5\% drop can be observed in the $8$-th epoch of \autoref{fig:curve} (in~\autoref{appendix:additional_results}) when switching from the pre-training to DEQ training.
The idea of Jacobian regularization~\cite{bai2021stabilizing} or phantom gradient~\cite{geng2021training} may alleviate this drop, but the gap, when compared to weight-tied ones, remains present as shown in \autoref{fig:curve}. \looseness=-1

\begin{figure}[!h]
	\centering
	\includegraphics[width=1.\textwidth]{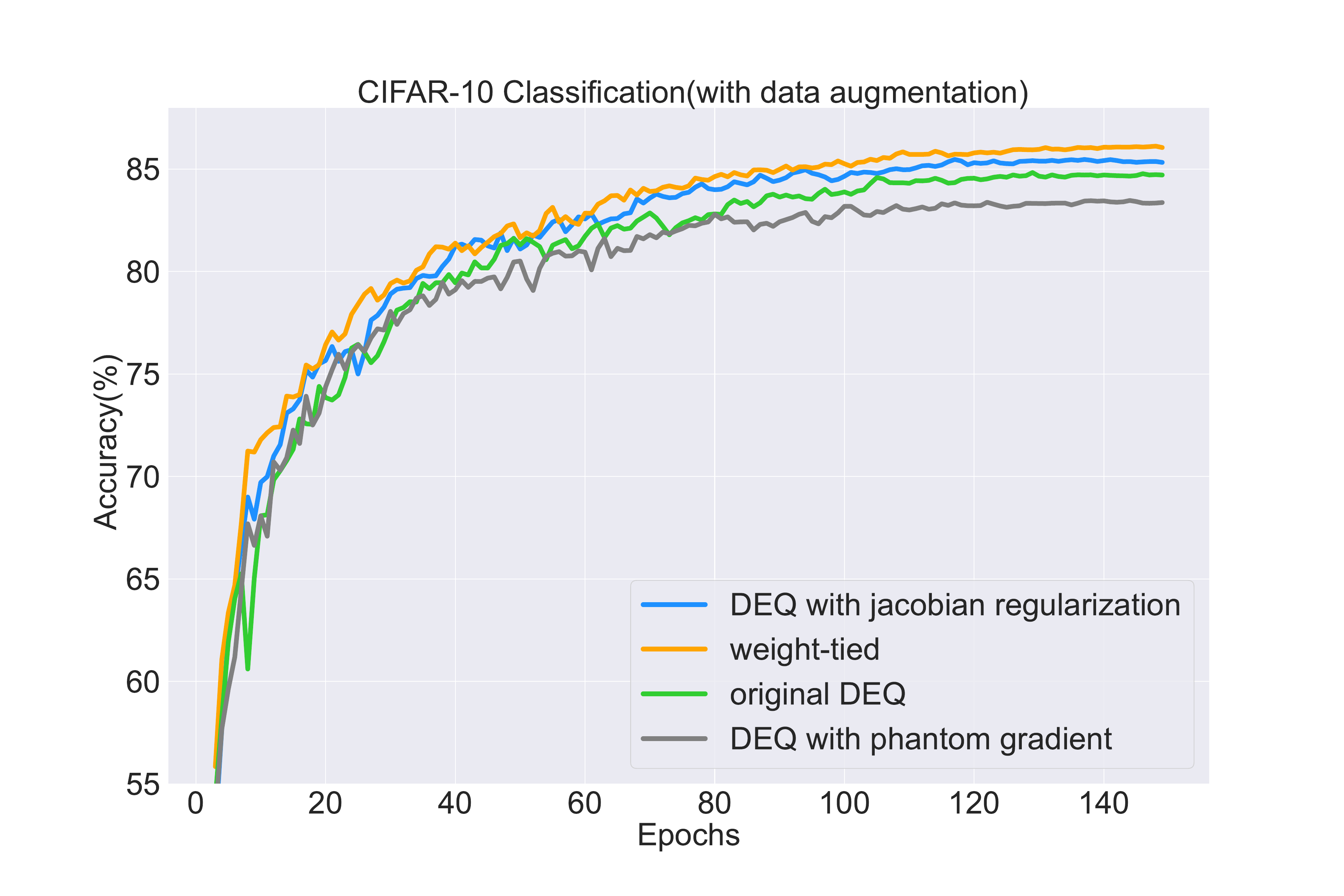}
	\vspace{-2.5em}
	\caption{\small
		\textbf{The ineffectiveness and instability of DEQ variants over weight-tied models on the training dynamics.} The test is conducted using the ResNet model for CIFAR-10 image classification (w/ data augmentation). We depict the test accuracy over 150 training epochs, where the performance of various DEQ variants consistently under-perform the simple weight-tied model. All DEQ variants exhibit instability during training while the weight-tied model training is more stable. For fair comparison, we use Adam optimizer for all methods and use the hyper-parameters as suggested in the open-source implementations of each method. SGD optimizer is used in~\cite{geng2021training} for the phantom gradient; however, no dramatic difference can be observed in our evaluation.
	}
	\label{fig:curve}
\end{figure}

\subsection{Interpreting Multi-Mask Weight-Tied Model}
\begin{figure}[!h]
	\centering
	\begin{minipage}{0.45\linewidth}
		\centering
		\includegraphics[width=0.9\linewidth]{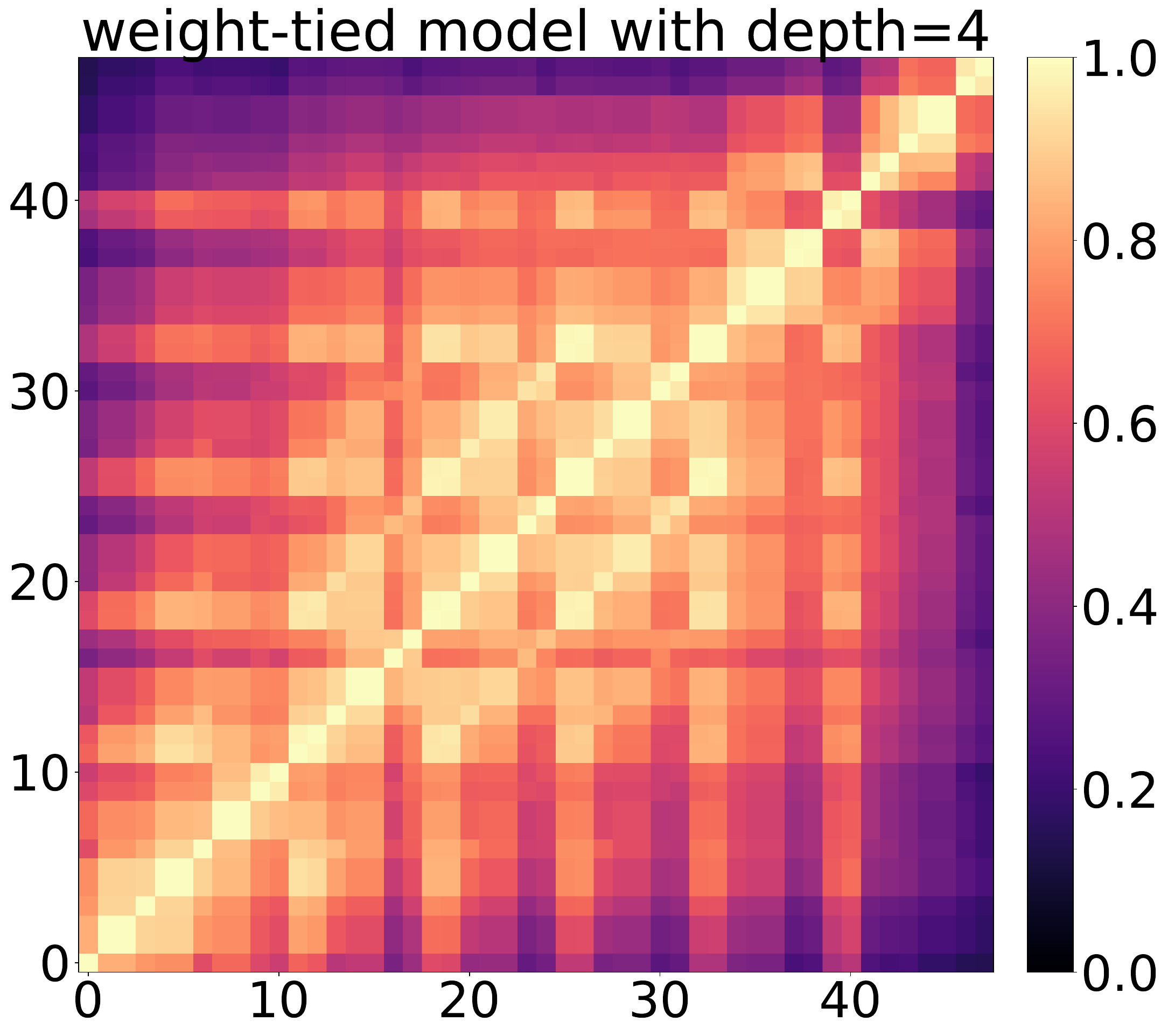}
	\end{minipage}
	\begin{minipage}{0.45\linewidth}
		\centering
		\includegraphics[width=0.9\linewidth]{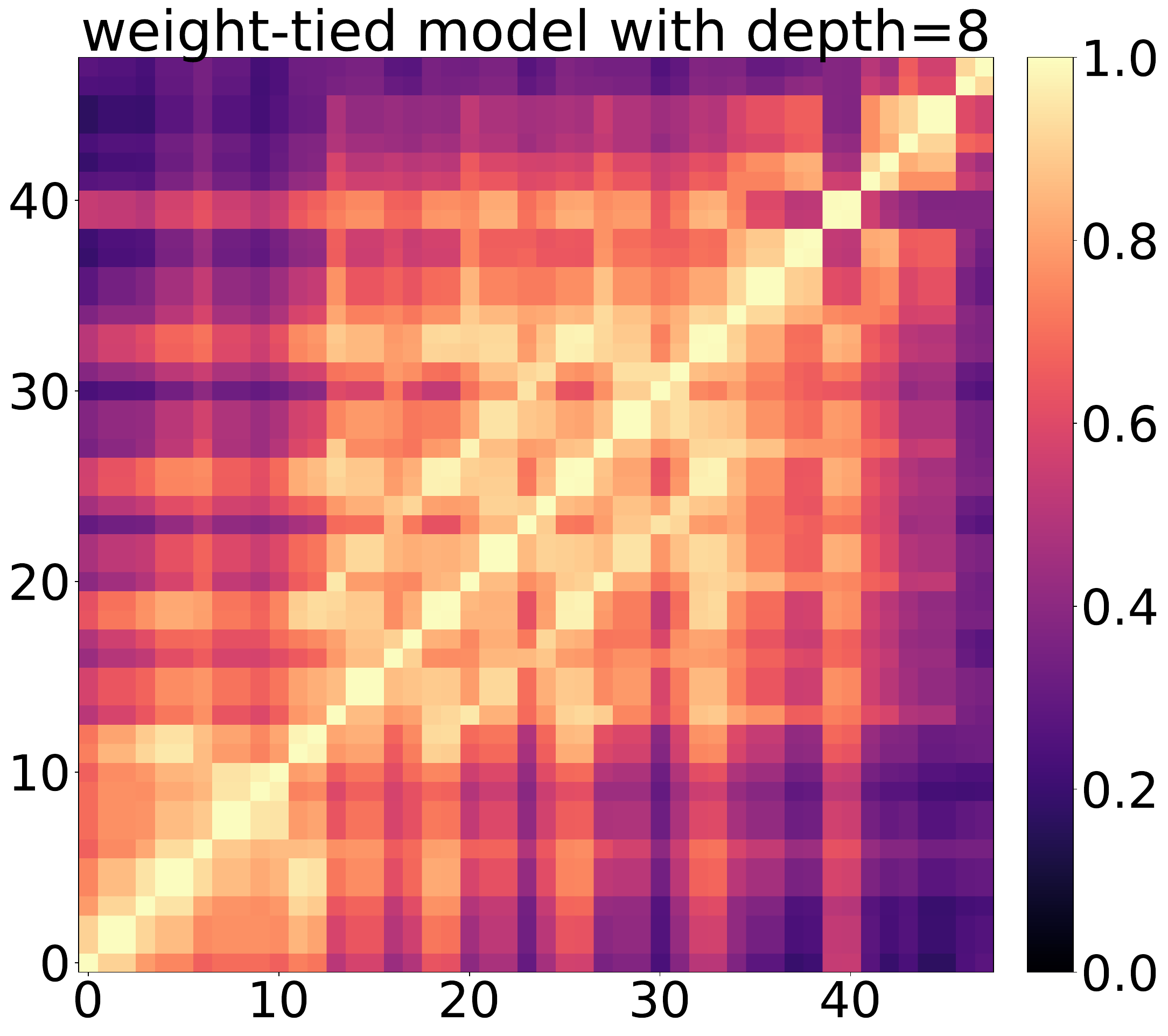}
	\end{minipage}
	\begin{minipage}{0.45\linewidth}
		\centering
		\includegraphics[width=0.9\linewidth]{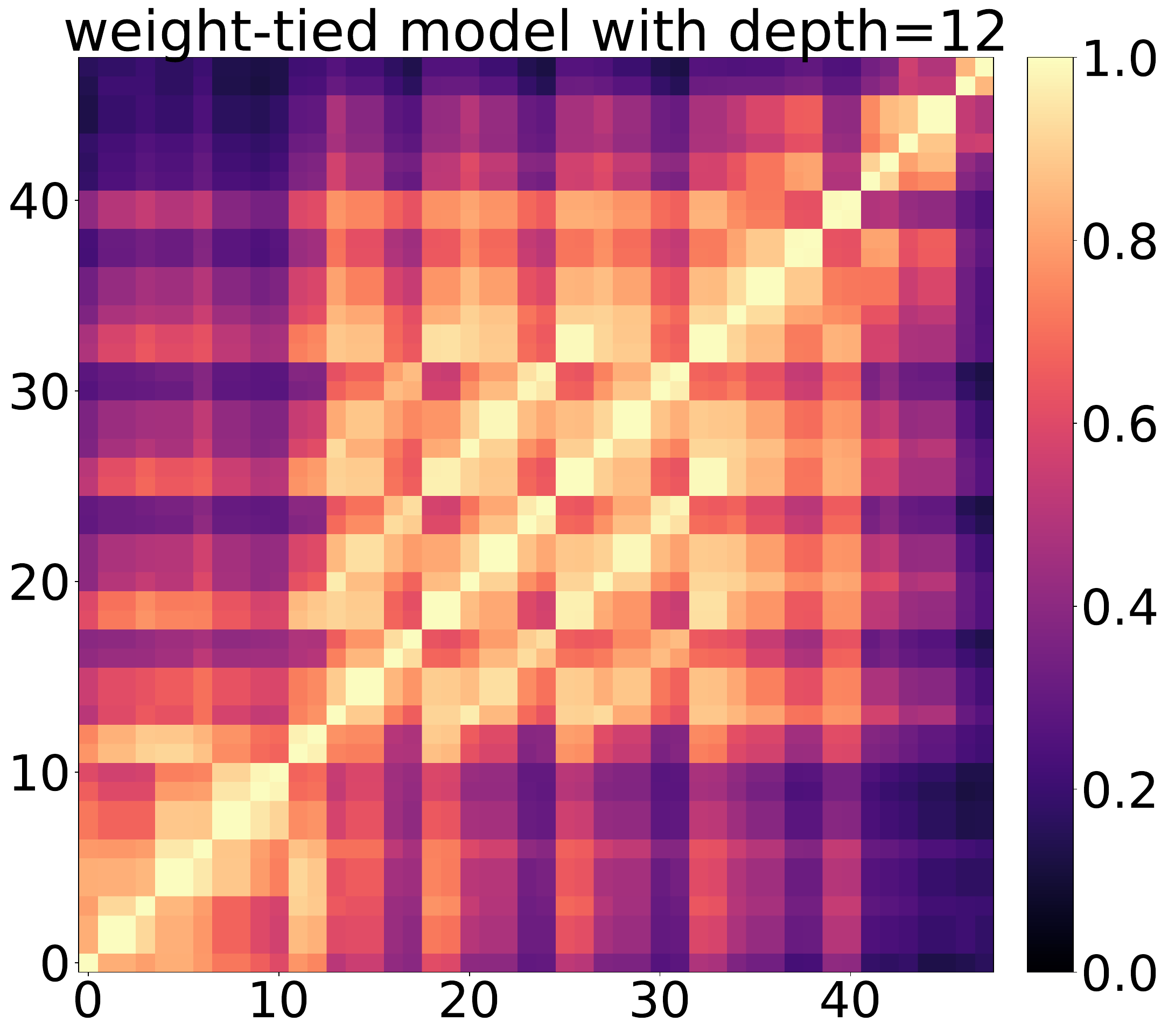}
	\end{minipage}
	\begin{minipage}{0.45\linewidth}
		\centering
		\includegraphics[width=0.9\linewidth]{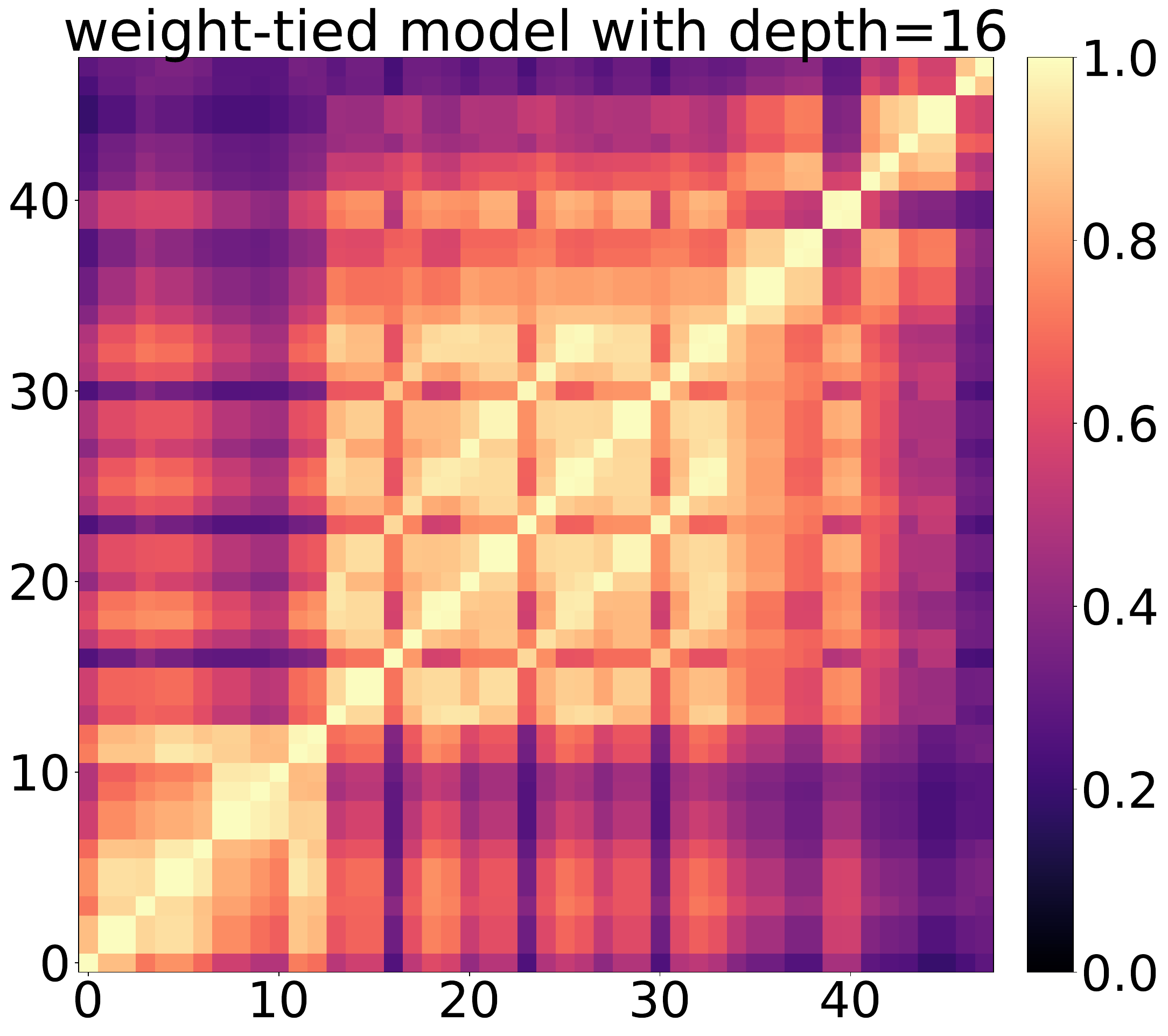}
	\end{minipage}
	\begin{minipage}{0.45\linewidth}
		\centering
		\includegraphics[width=0.9\linewidth]{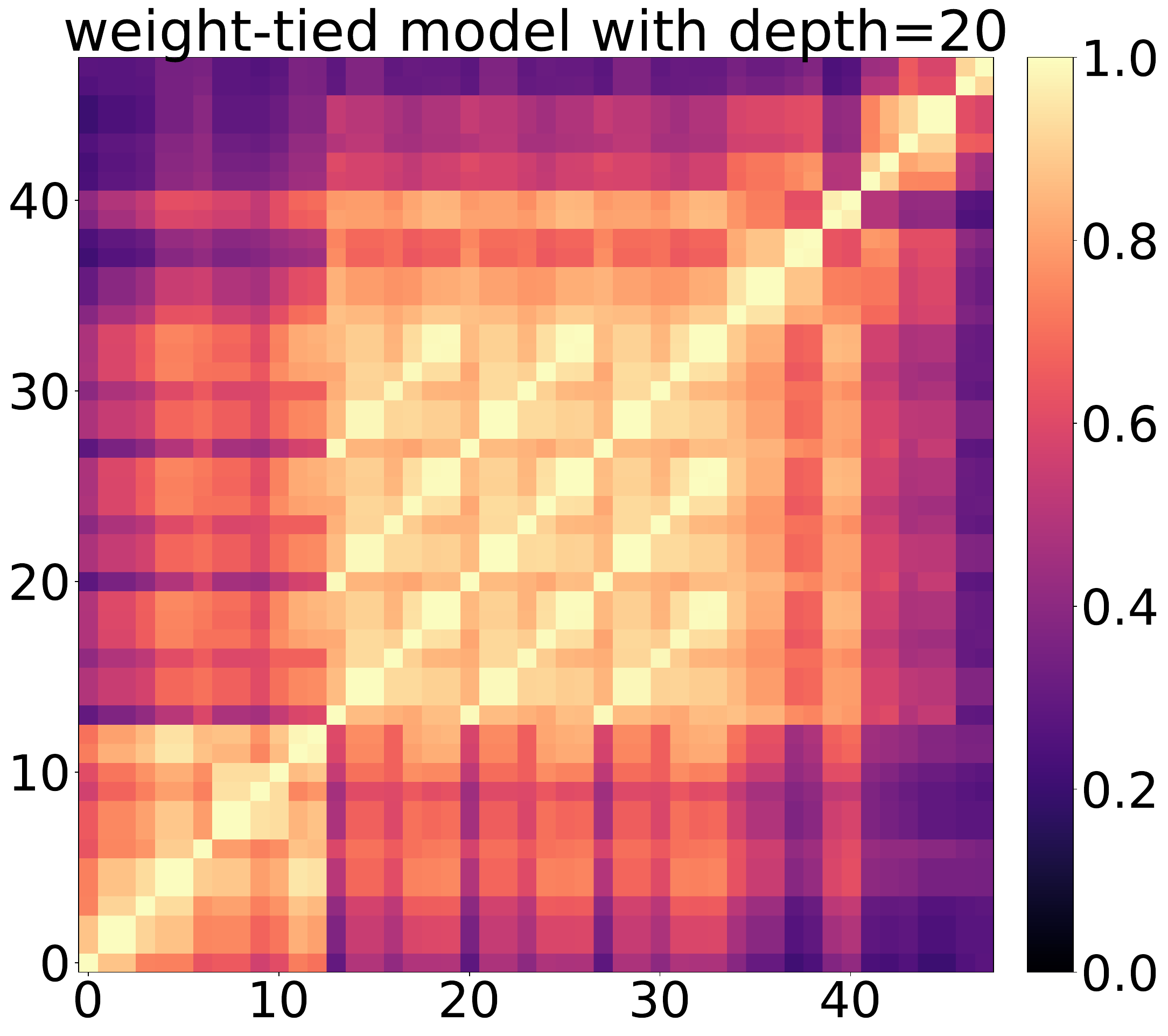}
	\end{minipage}
	\begin{minipage}{0.44\linewidth}
		\centering
		\includegraphics[width=1\linewidth]{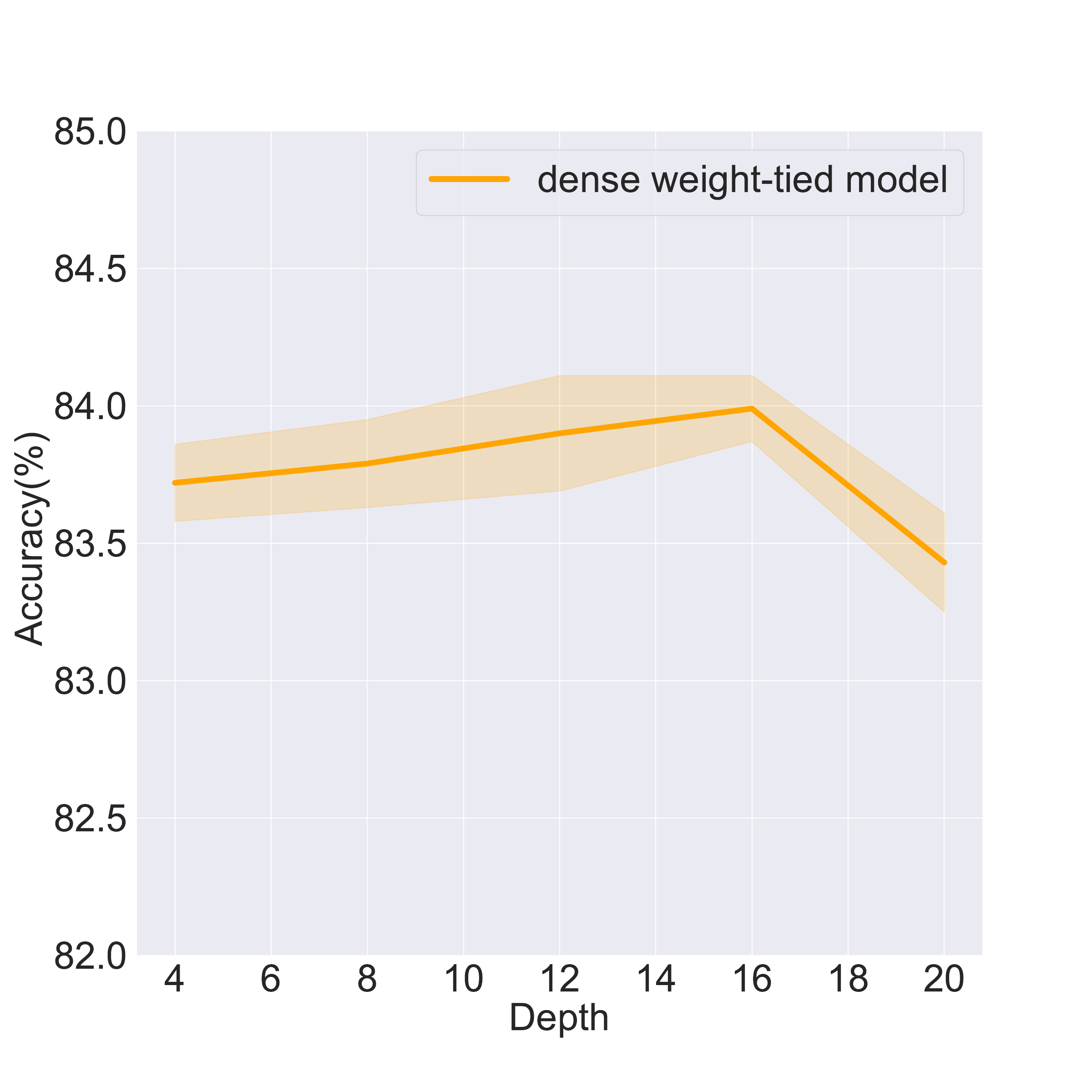}
	\end{minipage}
	\caption{\small
		\textbf{Larger depth strengthens high output similarity pattern.}
		The figure shows layer output similarity and performance of dense weight-tied model under varying depths of 4, 8, 12, 16 and 20. The square color in position (i,j) represents the output similarity between the i-th layer and the j-th layer. To demonstrate this pattern more clearly, we have selected 4 tied-layers from each model, spaced evenly.
	}
	\label{cka-depth}
\end{figure}

\endgroup

\end{document}